\documentclass[runningheads]{llncs}

\usepackage{booktabs} 

\usepackage{subfig}
\usepackage{graphicx,color}
\usepackage{colortbl}
\usepackage{algorithm}
\usepackage{algorithmic}
\usepackage{bm}
\usepackage{amsmath}
\usepackage{soul}
\usepackage{rotating}
\usepackage{multirow}
\usepackage{verbatim}
\usepackage[dvipsnames]{xcolor}

\usepackage{marvosym}
\usepackage{multiobjective}

\usepackage{todonotes}

\usepackage{ulem}

\usepackage{threeparttable}

\usepackage{enumitem}
\setlist{leftmargin=4mm}

\usepackage[T1]{fontenc}
\usepackage[usestackEOL]{stackengine}
\usepackage{lscape}

\usepackage{tabularx}

\newcounter{ALC@tempcntr}

\definecolor{mypink}{cmyk}{0, 0.7808, 0.4429, 0.1412}

\definecolor{blue-light}{RGB}{66, 191, 244}

\newcommand{\ignore}[1]{}

\begin{document}

\title{
MATE: A Model-based Algorithm Tuning Engine
}
\subtitle{A proof of concept towards transparent feature-dependent parameter tuning using symbolic regression}


\author{Mohamed El Yafrani\inst{1} \and
Marcella Scoczynski \inst{2} \and
Inkyung Sung\inst{1} \and
Markus Wagner\inst{3} \and
Carola Doerr\inst{4} \and
Peter Nielsen\inst{1}
}
\authorrunning{El Yafrani et al.}
\institute{
Operations Research group, Aalborg University, Denmark\\
\and
Federal University of Technology Paran\'a (UTFPR), Brazil\\
\and
Optimisation and Logistics Group, The University of Adelaide, Australia\\
\and
Sorbonne Universit\'e, CNRS, LIP6, Paris, France\\
}
\maketitle

\begin{abstract}
In this paper, we introduce a Model-based Algorithm Tuning Engine, namely MATE, where the parameters of an algorithm are represented as expressions of the features of a target optimisation problem. In contrast to most \textit{static} (feature-independent) algorithm tuning engines such as \textit{irace} and \textit{SPOT}, our approach aims to derive the best parameter configuration of a given algorithm for a specific problem, exploiting the relationships between the algorithm parameters and the features of the problem. 
We formulate the problem of finding the relationships between the parameters and the problem features as a symbolic regression problem and we use genetic programming to extract these expressions in a human-readable form. For the evaluation, we apply our approach to the configuration of the (1+1) EA and RLS algorithms for the OneMax, LeadingOnes, BinValue and Jump optimisation problems, where the theoretically optimal algorithm parameters to the problems are available as functions of the features of the problems. Our study shows that the found relationships typically comply 
with known theoretical results -- this 
demonstrates (1) the potential of model-based parameter tuning as an alternative to existing static algorithm tuning engines, and (2) its potential to discover relationships between algorithm performance and instance features in human-readable form.

\end{abstract}

\sloppy

\section{Motivation}
\label{sect:sect1}


The performance of many algorithms is highly dependent on tuned parameter configurations made with regards to the user's preferences or performance criteria~\cite{belkhir2016feature}, such as the quality of the solution obtained in a given CPU cost, the smallest CPU cost to reach a given solution quality, the probability to reach a given quality, with given thresholds, and so on. 
This configuration task can be considered as a second layer optimisation problem~\cite{hoos2012programming} relevant in the fields of optimisation, machine learning and AI in general. 
It is a field of study that is increasingly critical as the prevalence of the application of such methods is expanded. Over the years, a range of automatic parameter tuners have been proposed, thus leaving the configuration to a computer rather than manually searching for performance-optimised settings across a set of problem instances. These tuning environments can save time and achieve better results~\cite{ansotegui2009gender}.

\begin{figure}[!htbp]
\centering
\includegraphics[width=.8\textwidth]{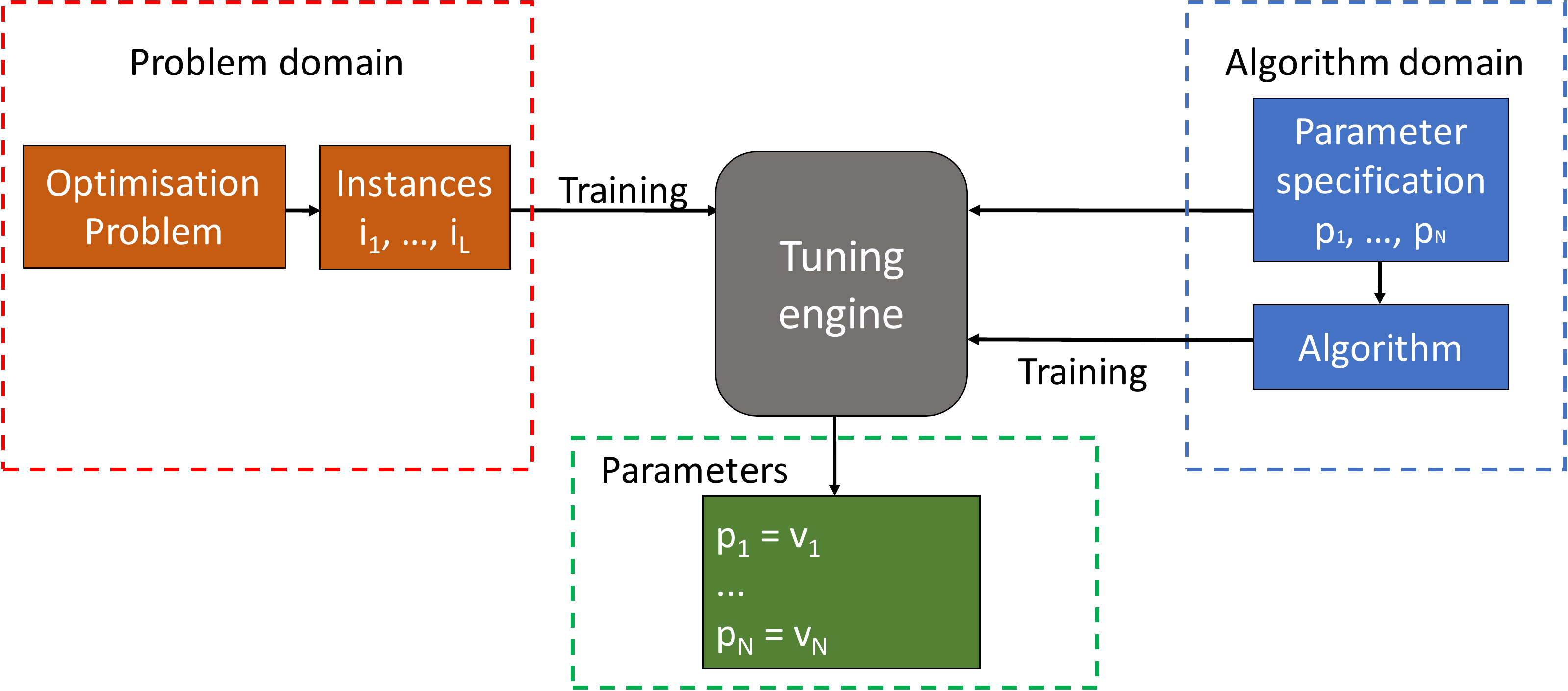}
\caption{Standard architecture of tuning frameworks.}
\label{fig:classic-tuning-engine}
\end{figure}

Among such automated algorithm configuration (AAC) tools, we cite
GGA~\cite{ansotegui2009gender}, ParamILS~\cite{hutter2009paramils}, SPOT~\cite{bartz2010spot} and irace~\cite{lopez2016irace}. These methods have been successfully applied to (pre-tuned) state-of-the-art solvers of various problem domains, such as mixed integer programming~\cite{hutter2010automated}, AI planning~\cite{fawcett2011fd}, machine learning~\cite{snoek2012practical}, or propositional satisfiability solving~\cite{hutter2017configurable}. Figure~\ref{fig:classic-tuning-engine} illustrates the abstract standard architecture adopted by these tools.

However, the outcomes of these tools are static (or feature-independent), which means an algorithm configuration derived by any of these tools is not changed depending on an instance of a target optimisation problem.
This leads to a significant issue as theoretical and empirical studies on various algorithms and problems have shown that parameters of an algorithm are highly dependent on features of a specific instance of a target problem~\cite{DoerrFastGA} such as the problem size~\cite{BottcherDN10,Witt13j}.


A possible solution to this issue is to cluster problem instances into multiple sub-groups by their size (and other potential features), then use curve fitting to map features to parameters~\cite{mascia2013tuning,yafrani2018efficiently}. A similar approach is also found in~\cite{liefooghe2017towards} that first partitions problem instances based the values of their landscape features and selects an appropriate configuration of a new problem instance based on its closeness to the partitions. However, the former approach does not scale well to multiple features and parameters, and the latter faces over-fitting issues due to the nature of the partitioning approach, making it difficult to assign an unseen instance to a specific group.


Some works have incorporated problem features in the parameter tuning process. SMAC~\cite{hutter2011sequential} and PIAC~\cite{leyton2002learning} are examples of model-based tools that consider instance features to define parameter values by applying machine learning techniques to build the model. However, an issue of these approaches is the low explainability of the outcome. For instance, while machine learning techniques such as random forest and neural networks can be used to map the parameters to problem features with a high accuracy, they are considered as black-boxes, i.e., the outcome is virtually impossible to understand or interpret. Explainability is an important concept, as not only it allows us to understand the relationships between input and output~\cite{rai2020explainable}, but in the context of parameter tuning, it can provide an outcome that can be used to inspire fundamental research~\cite{Friedrich2018heavyppsn,Friedrich2018heavygecco}.

To tackle these issues, we propose an offline algorithm tuning approach that extracts relationships between problem features and algorithm parameters using a genetic programming algorithm framework. We will refer to this approach as MATE, which stands for Model-based Algorithm Tuning Engine. The main contributions in this work are as follows:
\begin{enumerate}
  \item We formulate the model-based parameter tuning problem as a symbolic regression problem, 
  where knowledge about the problem is taken into account in the form of problem features;
  \item We implement an efficient Genetic Programming (GP) algorithm that configures parameters in terms of problem features; and
  \item In our empirical investigation, we rediscover asymptotically-correct theoretical results for two algorithms (1+1-EA and RLS) and four problems (OneMax, LeadingOnes, BinValue, and Jump). In these experiments, MATE shows its potential in algorithm parameter configuration to produce models based on instance features. 
\end{enumerate}

\section{Background}
\label{sec:sect2}


Several methods have tried to tackle the dependence between the problem features and the algorithm parameters.
The Per Instance Algorithm Configuration (PIAC)~\cite{leyton2002learning}, for example, can learn a mapping between features and best parameter configuration, building an Empirical Performance Model (EPM) that predicts the performance of the algorithm for sample (instance, algorithm/configuration) pairs. PIAC methodology has been applied to several combinatorial 
problems~\cite{hutter2006performance,xu2008satzilla,hutter2014algorithm} and continuous domains~\cite{Belkhir:2017:PIA:3071178.3071343}.

Sequential Model-based Algorithm Configuration (SMAC)~\cite{hutter2011sequential} is also an automated algorithm configuration tool which considers a model, usually a random forest, to design the relationship between a performance metric (e.g. the algorithm runtime) and algorithm parameter values. SMAC can also include problem features within the tuning process as a subset of input variables. 

Table \ref{tab:works} presents a summary for some state-of-the-art methods including the approach proposed in this paper. 
The term `feature-independent' means that the corresponding approach does not consider instance features.
`Model-based' approaches use a trained model (e.g. machine learning, regression, etc.) to design parameter configurations. 
Model-free approaches generally rely on an experimental design methodology or optimisation method to find parameter settings of an algorithm that optimise a cost metric on a given instance set.

\begin{table}[t]
\scriptsize
\centering
\caption{Summary of the state-of-the-art related works}
\label{tab:works}
\begin{tabular}{m{1.7cm}m{4.4cm}m{4cm}m{0.8cm}}
\toprule
\bf{Approach Name} & \bf Algorithm &\bf Characteristics & \bf Ref.\\
\midrule
\bf GGA	     & Genetic Algorithm                                                    & Feature-independent, model-free  & \cite{ansotegui2009gender} \\ \hline
\bf ParamILS & Iterated Local Search                                                & Feature-independent, model-free  & \cite{hutter2009paramils}  \\ \hline
\bf irace 	 & Racing procedure                                                     & Feature-independent, model-free  & \cite{lopez2016irace}      \\ \hline
\bf SPOT 	 & classical regression, tree-based, random forest and Gaussian process & Feature-independent, model-based & \cite{bartz2010spot}       \\ \hline
\bf PIAC 	 & Regression methods                                                   & Feature-dependent, model-based   & \cite{leyton2002learning}  \\ \hline
\bf SMAC 	 & Random Forest                                                        & Feature-dependent, model-based   & \cite{hutter2011sequential}\\ \hline
\bf MATE 	 & Genetic Programming                                                  & Feature-dependent, model-based, explainable &  \\ 
\bottomrule
\end{tabular}
\end{table}

The main differences between MATE and the other related approaches are: 
\begin{enumerate}
\item A transparent machine learning method (GP) is utilised to enable human-readable configurations (in contrast to, e.g., random forests, neural networks, etc.). 
\item The training phase is done on one specific algorithm and one specific problem in our approach -- the model is less instance-focused but more problem-domain focused by abstracting via the use of features. For example, the AAC experiments behind  \cite{Friedrich2018heavyppsn,Friedrich2018heavygecco} have guided the creation of new heavy-tailed mutation operators that were beating the state-of-the-art. Similarly, the AAC and PIAC experiments in \cite{Treude2019topicmodelling} 
showed model dependencies on easily-deducible instance features.
\end{enumerate}

Lastly, our present paper is much aligned with the recently founded research field ``Data Mining Algorithms Using/Used-by Optimisers (DUO)''~\cite{agrawal2018better}. There, data miners can generate models explored by optimisers; and optimisers can adjust the control parameters of a data miner.

\section{The MATE Framework}
\label{sec:sect3}

\subsection{Problem Formulation and Notation}

Let us denote an optimisation problem by $\mathcal{B}$ whose instances are characterised by the problem-specific features $\mathcal{F}=\{f_1, \dots, f_M\}$. A target algorithm $\mathcal{A}$ with its parameters $\mathcal{P}=\{p_1, \dots, p_N\}$ is given to address the problem $\mathcal{B}$. A set of instances $\mathcal{I}=\{i_1, \dots, i_L\}$ of the problem $\mathcal{B}$ and a $L \times M$ matrix $\mathcal{V}$, whose element value $v_{i,j}$ represents the $j$th feature value of the $i$th problem instance, are given.

Under this setting, we define the model-based parameter tuning problem as the problem of deriving a list of mappings $\mathcal{M}=\{m_1,\dots,m_N\}$ where each mapping $m_j \colon \mathbb{R}^M \to \mathbb{R}$, which we will refer to as a \textit{parameter expression}, returns a value for the parameter $p_j$ given feature values of an instance of the problem $\mathcal{B}$. 
Specifically, the objective of the problem is to find a parameter expression set $\mathcal{M^*}$, such that the performance of the algorithm $\mathcal{A}$ across all the problem instances in $\mathcal{I}$ is optimised.

\subsection{Architecture Overview}

\begin{figure}[!htbp]
\centering
\includegraphics[width=.8\textwidth]{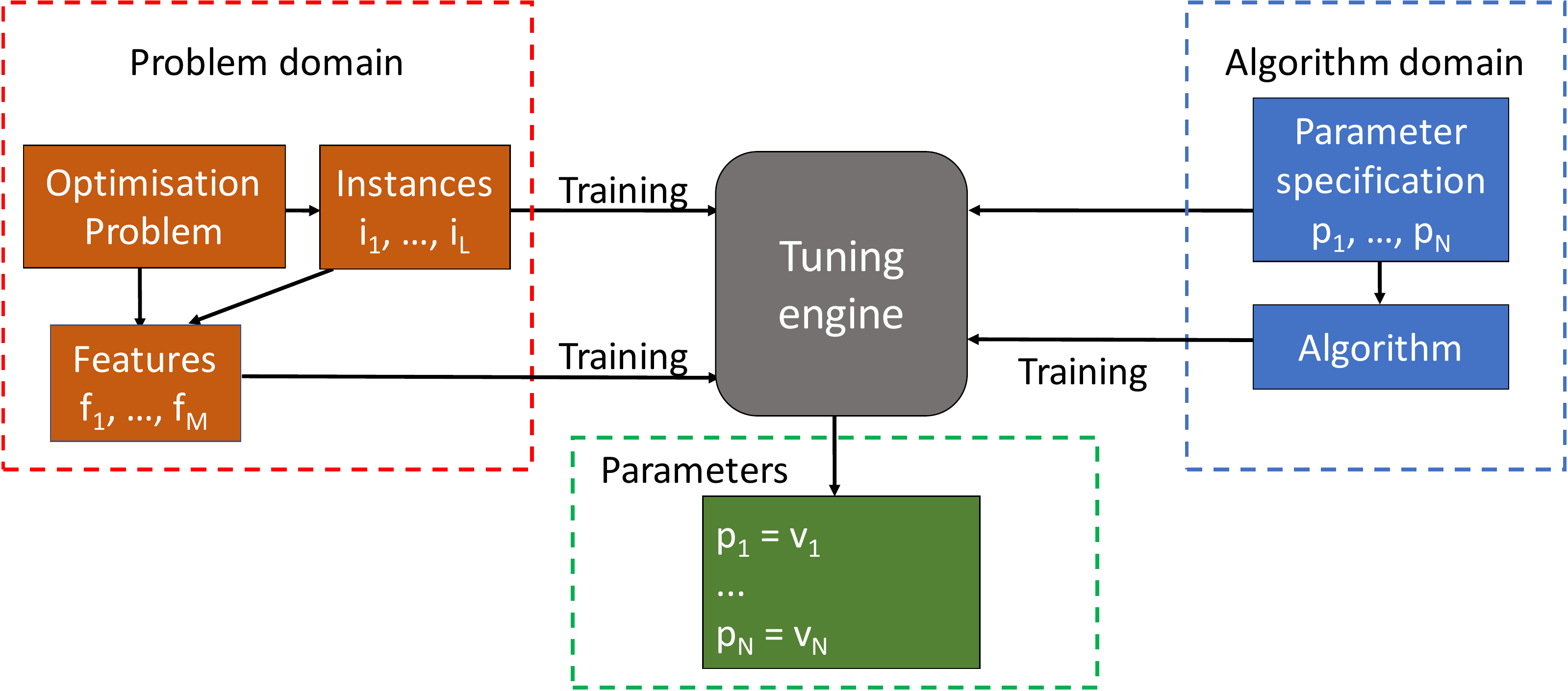}
\caption{Architecture of the proposed MATE framework}
\label{fig:mate-framework}
\end{figure}

In this section, we introduce our approach for parameter tuning based on the problem features. Figure~\ref{fig:mate-framework} illustrates the architecture of the MATE tuning engine. In contrast to static methods, we consider the features of the problem.
These feature are to be used in the training phase in addition to the instances, the target algorithm and the parameter specifications. Once the training is finished, the model can be used on unseen instances to find the parameters of the algorithm in terms of the problem feature values of the instance. 

For example, a desired outcome of applying the MATE framework can be:
\begin{itemize}
    \item Mutation probability of an evolutionary algorithm in terms of the problem size;
    \item Perturbation strength in an iterated local search algorithm in terms of the ruggedness of the instance and the problem size; and
    \item Population size of an evolutionary algorithm in terms of the problem size.
\end{itemize}

Note that all the examples include the problem size as a problem feature. In both theory and practice, the problem size is among the most important problem features, and it is usually known prior to the optimisation, without any need for a pre-processing step. More importantly, an extensive number of theoretical studies showed that the optimal choice of parameters is usually expressed in terms of the problem size (see, e.g.~\cite{BottcherDN10,DoerrFastGA,Witt13j}).

\subsection{The Tuning Algorithm}

We use a tree-based Genetic Programming system as the tuning algorithm. It starts with a random population of trees, where each tree represents a potential parameter expression. 
Without loss of generality, we assume that the target problem is always a maximisation problem\footnote{The current MATE implementation is publicly available at \url{https://gitlab.com/yafrani/mate}}.

\subsubsection{The Score Function and Bias Reduction}

The score function is expressed as the weighted sum of the obtained objective values on each instance in the training set $\mathcal{I}$. Using the notations previously introduced, the score is defined in Equation~\eqref{eq:score}:

\begin{equation}\label{eq:score}
S(t) = \frac{1}{L} \Sigma_{i \in I} \frac{ z_{\mathcal{A}} (m_1(v_{i,1},\dots,v_{i,M}), \dots , m_N(v_{i,1},\dots,v_{i,M}), i) }{R_i}
\end{equation}
where:
\begin{itemize}
    \item $S(.)$ is the GP score function,
    \item $z_{\mathcal{A}}(\varphi_1,\dots,\varphi_N, i)$ is a function measuring the goodness of applying the algorithm $A$ with the parameter values $\varphi_1, \dots, \varphi_N$ to instance $i$,
    \item $R_i$ is the best known objective value for instance $i$.
\end{itemize}


The weights are used as a form of normalisation to reduce the bias some instances might induce. 
A solution to address this issue would be to use the optimal value or a tight upper bound. However, since we assume the such values are unknown (the problem itself can be unknown), we use the best known objective value ($R_i$) as a reference instead. In order to always ensure that score is well contained, the reference values are constantly updated whenever possible during the tuning process.

\subsubsection{Replacement Strategy -- Statistical Significance and Bloat Control}

As the target algorithm can be stochastic, it is mandatory to perform multiple runs to ensure statistical significance (refer to Table~\ref{tab:mate-setup}). Thus, the replacement of trees is done based on the Wilcoxon rank-sum test.

Another aspect to take into account during the replacement process is bloat control. In our implementation, we use a simple bloat minimisation method based on the size of tree (number of nodes). 

Given a newly generated tree ($Y$), we compare it against each tree ($X$) in the current population starting from the ones with the lowest scores using the following rules:
\begin{itemize}
  \item If $Y$ is deemed to be significantly better than $X$ (using the Wilcoxon test).
  then we replace $X$ with $Y$ irrespective of the sizes.
  \item If there is no statistical significance between $X$ and $Y$, but $Y$ has a smaller size than $X$, then we replace $X$ with $Y$.
  \item Otherwise, we do not perform the replacement.
\end{itemize}

\section{Computational Study}
\label{sec:sect4}

\subsection{Experimental Setting}
To evaluate our framework, we consider two target algorithms, the (1+1)~EA($\mu$) and RLS($k$). The (1+1)~EA($\mu$) is a simple hill-climber which uses standard bit mutation with mutation rate $\mu$. RLS($k$) differs from the (1+1)~EA($\mu$) only in that it uses the mutation operator that always flips $k$ uniformly chosen, pairwise different bits. That is, the mutation strength $k$ is deterministic in RLS, whereas it is binomially distributed in case of the (1+1)~EA($\mu$), Bin$(n,\mu)$, where $n$ is the number of bits.


We use MATE to configure the two algorithms for the four different problems with different time budgets as summarised in Table~\ref{tab:problems}. In the table, the features of the problems used to tune the algorithm parameters and the different feature values chosen to generate problem instances of the problems are also presented. These problems have been chosen because they are among the best studied benchmark problems in the theory of evolutionary algorithms~\cite{DoerrN20}. The details of our GP implementation for the experiments are presented in Table~\ref{tab:mate-setup}.
Based on Table \ref{tab:mate-setup} and the set of features, our GP method uses a minimalistic set of $6$ terminals at most: $m$, $n$ and $\{1,2,-1,-2\}$. 

\begin{table}[t]
\centering
\scriptsize
\caption{Summary of problems\label{tab:problems}}
\label{tab:problems}
\begin{tabular}{m{2.6cm}|m{3.5cm}|m{5.5cm}}
\toprule
\bf{Problem} & \bf{Features} & \bf{Training set}\\
\midrule
\bf OneMax($n$) 	 &$n$: number of bits & $n=10,20,50,100,200,500$ \\ \hline
\bf BinValue($n$) 	 &$n$: number of bits & $n=10,20,50,100,200,500$ \\ \hline
\bf LeadingOnes($n$) &$n$: number of bits & $n=10,20,50,100,200,500$ \\ \hline
\bf Jump($m,n$)&
$m$: width of region with bad fitness values\newline
$n$: number of bits & 
$(m,n)=(2,10), (3,10), (4,10), (5,10),\newline
(2,20),(3,20), (4,20),\newline
(2,50),(3,50),\newline
(2,100),(3,100),\newline
(2,200)$
\\ \bottomrule
\end{tabular}
\end{table}

\begin{table}[t]
\centering
\scriptsize
\caption{MATE setup\label{tab:mate-setup}}
\begin{tabular}{m{7cm}|m{3.5cm}}
\toprule
\bf{Attribute/Parameter} & \bf{Value/Content}\\ 
\midrule
Terminals & $\{1,2,-1,-2\} \bigcup \mathcal{F}$\\ \hline
Functions & Arithmetic operators\\ \hline
Number of GP generations & $100$ \\ \hline
Population size & $20$ \\ \hline
Tournament size & $5$ \\ \hline
Replacement rate & $<75\%$ \\ \hline
Initialisation & grow ($50\%$) and full ($50\%$) methods \\ \hline
Mutation operator & random mutations \\ \hline
Mutation probability & $0.2$ \\ \hline
Crossover operator & sub-tree gluing \\ \hline
Crossover rate & $80\%$ \\ \hline
Number of independent runs of target algorithm & $10$ \\ \hline
$p$-value for the Wilcoxon ranksum test & $0.02$ \\ 
\bottomrule
\end{tabular}
\end{table}

It is worth noting that we are focusing in this paper on tuning algorithms with a single parameter. This is done to deliver a first prototype that is validated on algorithms and problems extensively studied by the EA theory community. An extension to tuning several algorithm parameters forms an important direction for future work. 

For example, given a budget of $(1+o(1)) e n \ln(n)$, it is known that the (1+1)EA$(1/n)$ optimises the OneMax function as well as any other linear functions with a decent probability. It is also known that the $1/n$ is asymptotically optimal~\cite{Lengler15}. Note, though, that such \textit{fixed-budget results} are still very sparse~\cite{Jansen20}, since the theory of EA community largely focuses on expected optimisation times. Since these can nevertheless give some insight into the optimal parameter settings, we note the following: 

\begin{itemize}
    \item OneMax and BinValue: the (1+1)EA$(1/n)$ optimises every linear function in expected time $e n \ln(n)$, and no parameter configuration has smaller expected running time, apart from possible lower order terms~\cite{Witt13j}. For RLS, it is not difficult to see that $k=1$ yields an expected optimisation time of $(1+o(1)) n \ln(n)$, and that this is the optimal (static) mutation strength;
    \item LeadingOnes: on average, RLS(1) needs $n^2/2$ steps to optimise LeadingOnes. This choise also minimises the expected optimisation time. For the (1+1)~EA, $\mu\approx 1.59/n$ minimises the expected optimisation time, which is around $0.77 n^2$ for this setting~\cite{BottcherDN10}. The standard mutation rate $\mu=1/n$ requires $0.86 n^2$ evaluations, on average, to locate the optimum, of the LeadingOnes function. For LeadingOnes, it is known that the optimal parameter setting drastically depends on the available budget. This can be inferred from the proofs in~\cite{BottcherDN10,Doerr19domi}; and
    \item Jump: mutation rate $m/n$ minimises the expected optimisation time of the (1+1)~EA on Jump$(m,n)$, which is nevertheless $\Theta((e/m)^m n^m)$~\cite{DoerrFastGA}. 
\end{itemize}

\subsection{Performance Analysis}

\subsubsection{Training Phase}
The experimental study is conducted by running MATE ten times on each algorithm, problem and budget combination (refer to Table \ref{tab:results-boxplots} for the list of budgets).
This results in an elite population of $20$ individuals for each setting, from which we select the top $5$ expressions in terms of the score. These results are then merged and the $3$ most frequent expressions are selected. 
For instance, the expression $2/n$ for OneMax with $0.5enln(n)$ appears 92 times over the 200 individuals (population size (20) $\times$ runs (10)). 

In the current implementation, expression types (integers and non-integers) are not taken into account during the evolution. Therefore, the resulting expressions are converted into integers in the case of RLS by merging all real numbers $r$ using $\lfloor r \rfloor$ (e.g. $k=3/2$ will be replaced by $k=1$). On the other hand, expressions are simplified for EA by eliminating additive constants (e.g. $\mu=1/(n+1)$ is replaced by $\mu=1/n$).


\subsubsection{Evaluation Phase I}
To assess the performance of MATE, we evaluate for each problem-budget combination each of the top $3$ most frequent expressions, by running them $100$ independent times on each training dimension. We then normalise the outputs as in Equation~\eqref{eq:score}. The results are shown in the box plots in Table~\ref{tab:results-boxplots}.

\begin{table}[!htbp]\scriptsize
\centering
\caption{Results for $20$ settings.
}
\label{tab:results-boxplots}
\renewcommand{\arraystretch}{1.0}
\begin{threeparttable}
\begin{tabular}{|m{.25cm}|l|l|l|l|}
\hline
  & \multicolumn{2}{l|}{\textbf{1+1-EA}}  & \multicolumn{2}{l|}{\textbf{RLS}}  \\ \hline
  & \textbf{Budget}   & \textbf{Result}   & \textbf{Budget} & \textbf{Result}  \\ \hline

\multirow{3}{*}{\rotatebox[origin=c]{90}{\bf OneMax}}
  & $0.5en\ln(n)$ & \includegraphics[width=.4\textwidth]{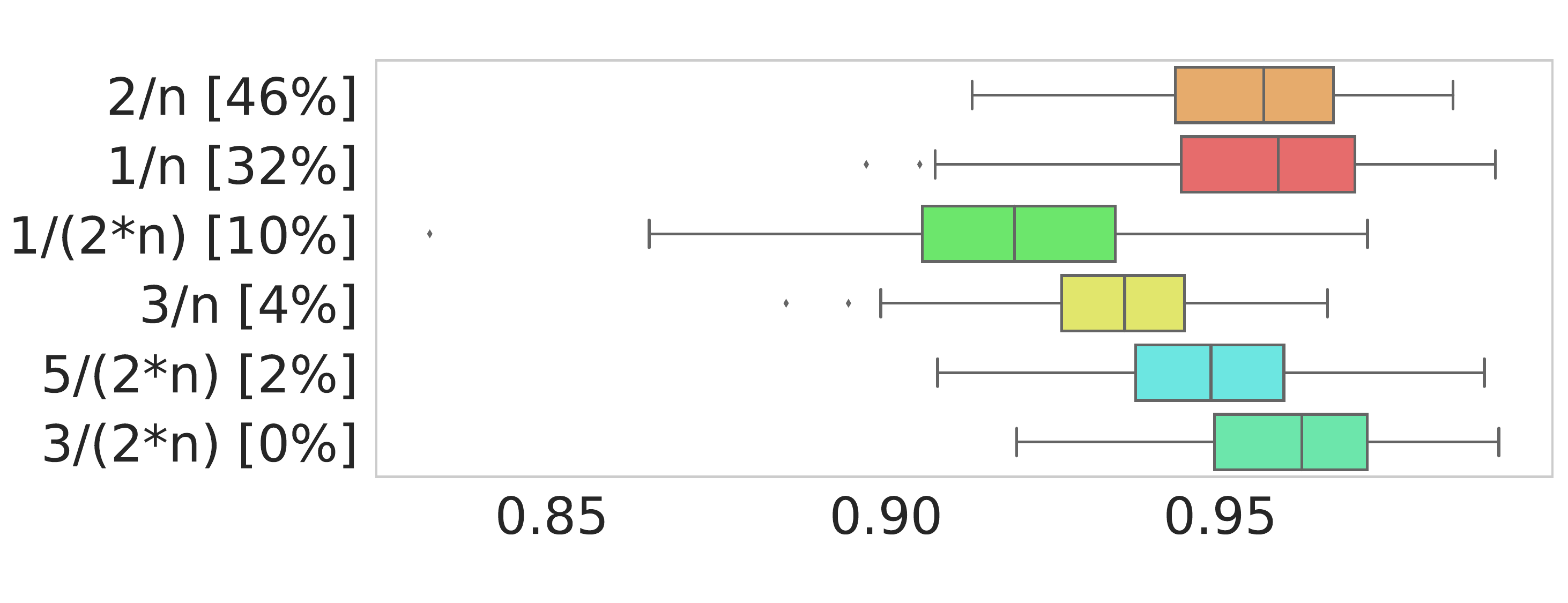}  & $n\ln(n)$*      & \includegraphics[width=.35\textwidth,clip,trim=70 0 0 0]{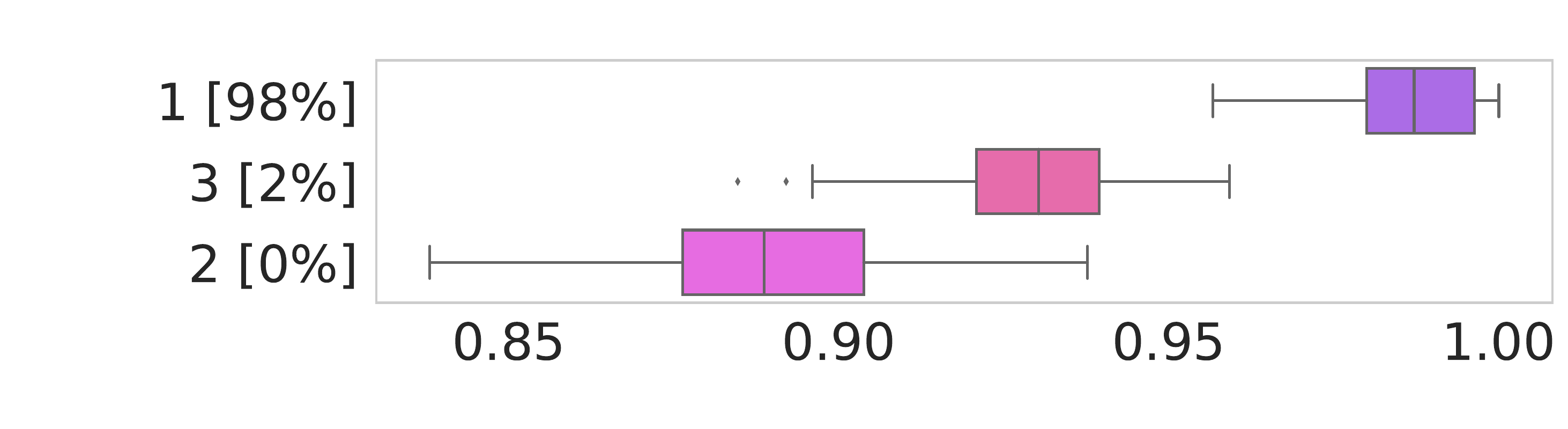}  \\ \cline{2-5}
  & $  e n\ln(n)$* & \includegraphics[width=.4\textwidth]{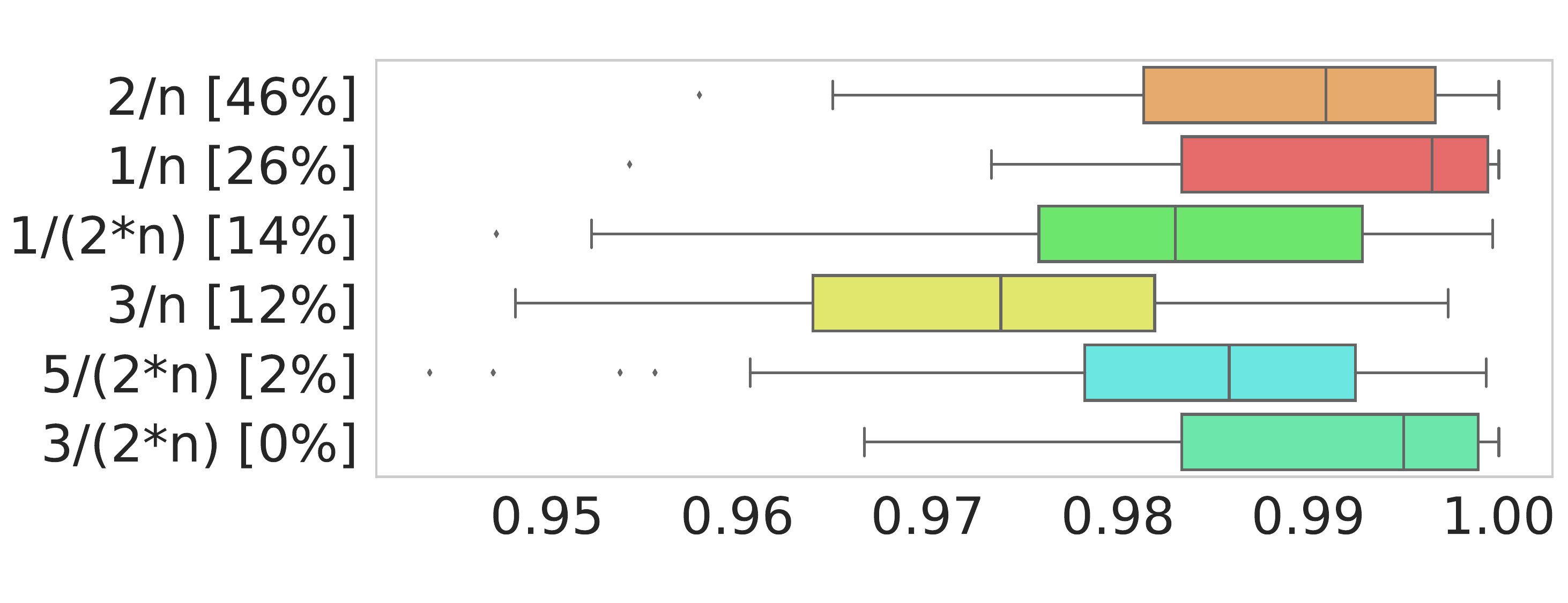}  & $2n\ln(n)$**     & \includegraphics[width=.35\textwidth,clip,trim=70 0 0 0]{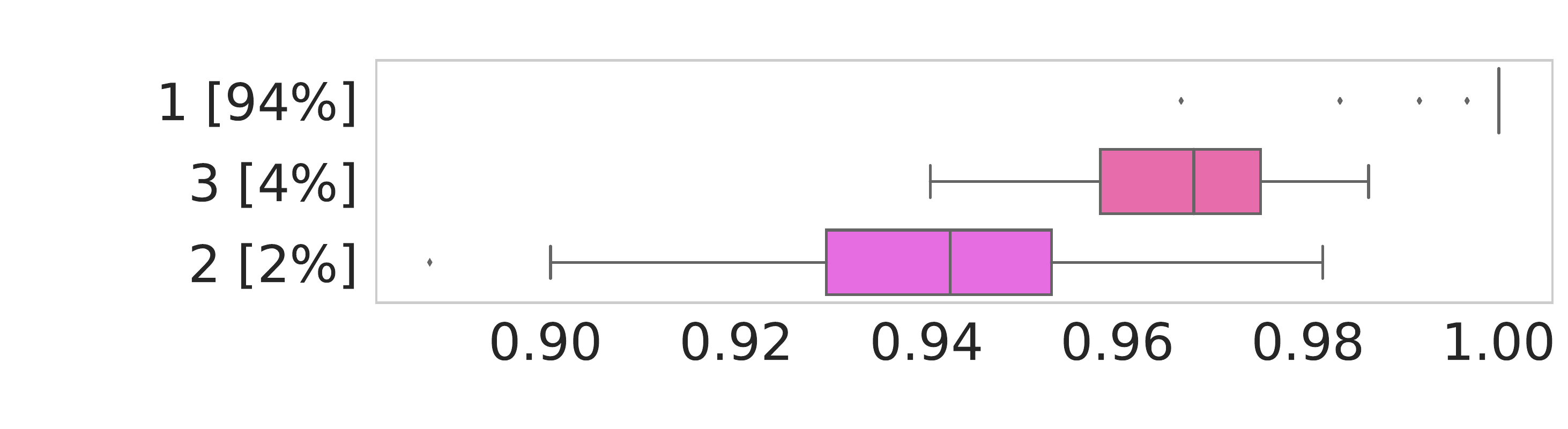}  \\ \cline{2-5}
  & $2 e n\ln(n)$** & \includegraphics[width=.4\textwidth]{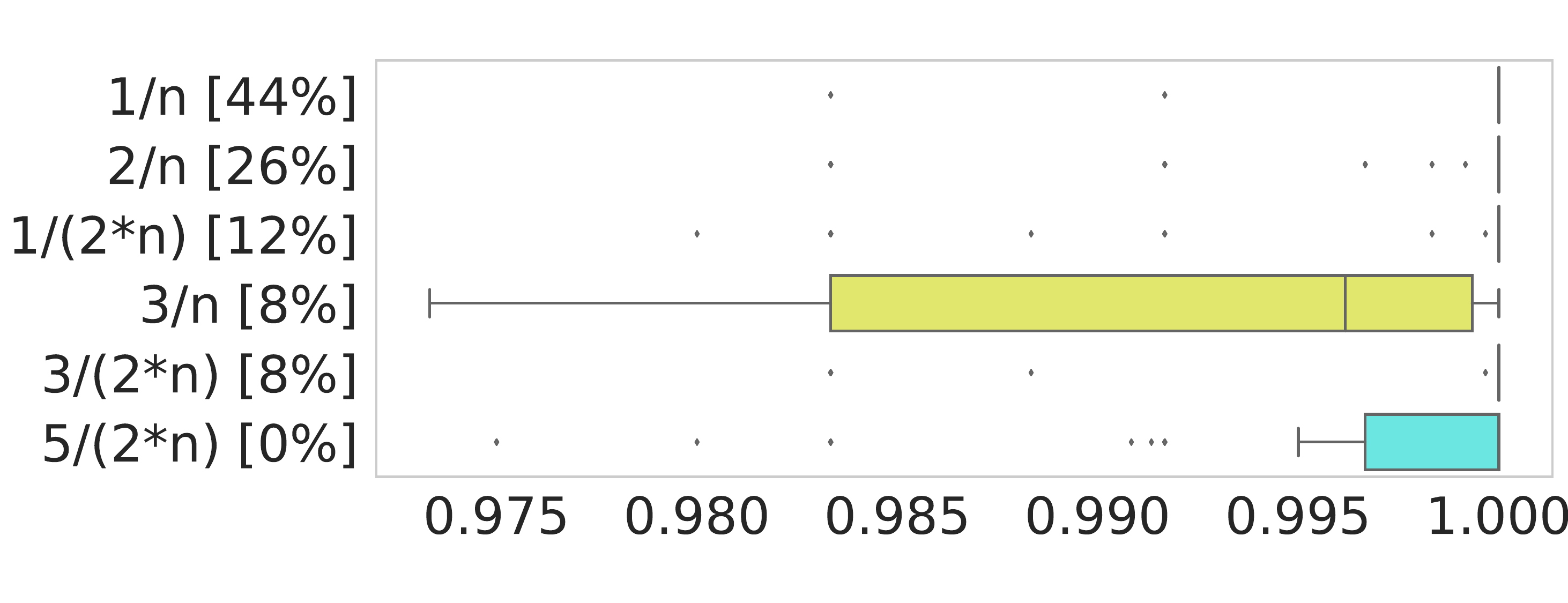}  &                &                                                              \\ \hline
\multirow{3}{*}{\rotatebox[origin=c]{90}{\bf BinValue}}
  & $0.5en\ln(n)$ & \includegraphics[width=.4\textwidth]{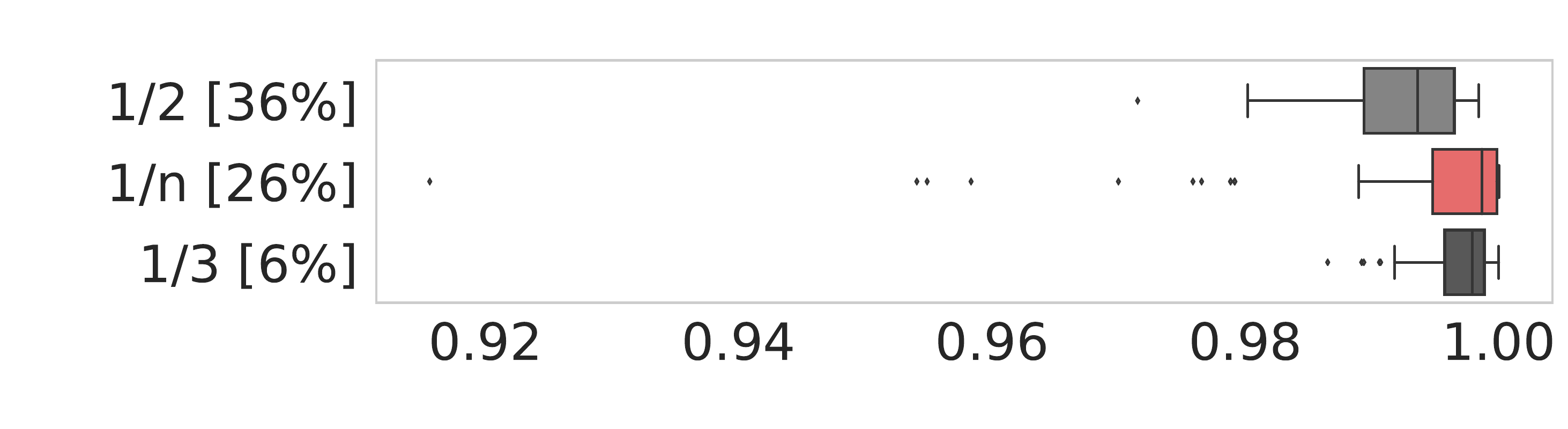}  & $0.5 n\ln(n)$  & \includegraphics[width=.35\textwidth,clip,trim=70 0 0 0]{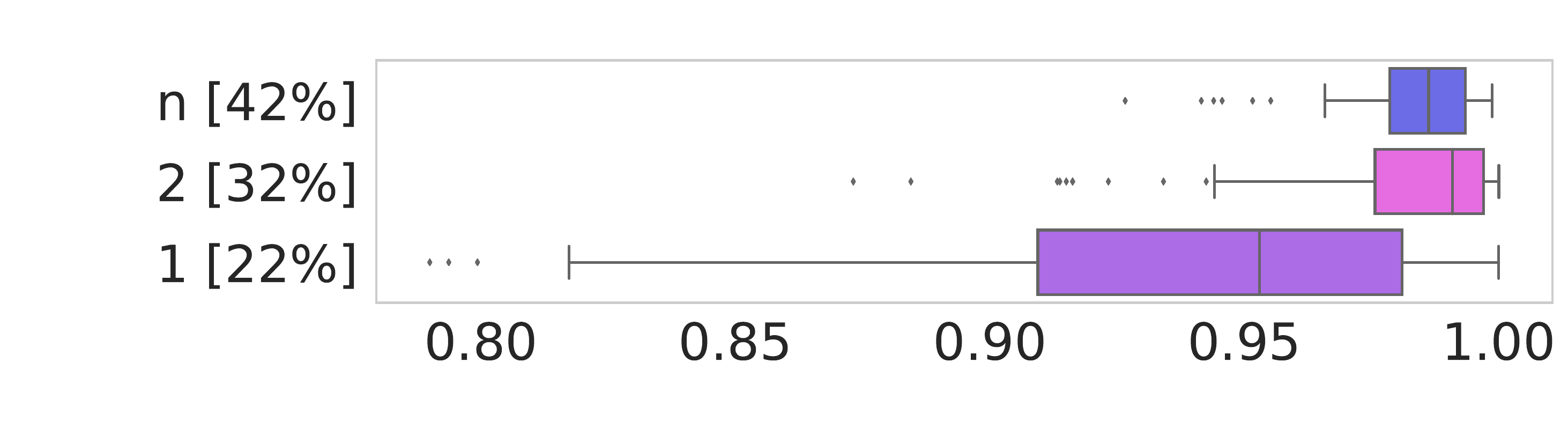}  \\ \cline{2-5}
  & $e n\ln(n)$*   & \includegraphics[width=.4\textwidth]{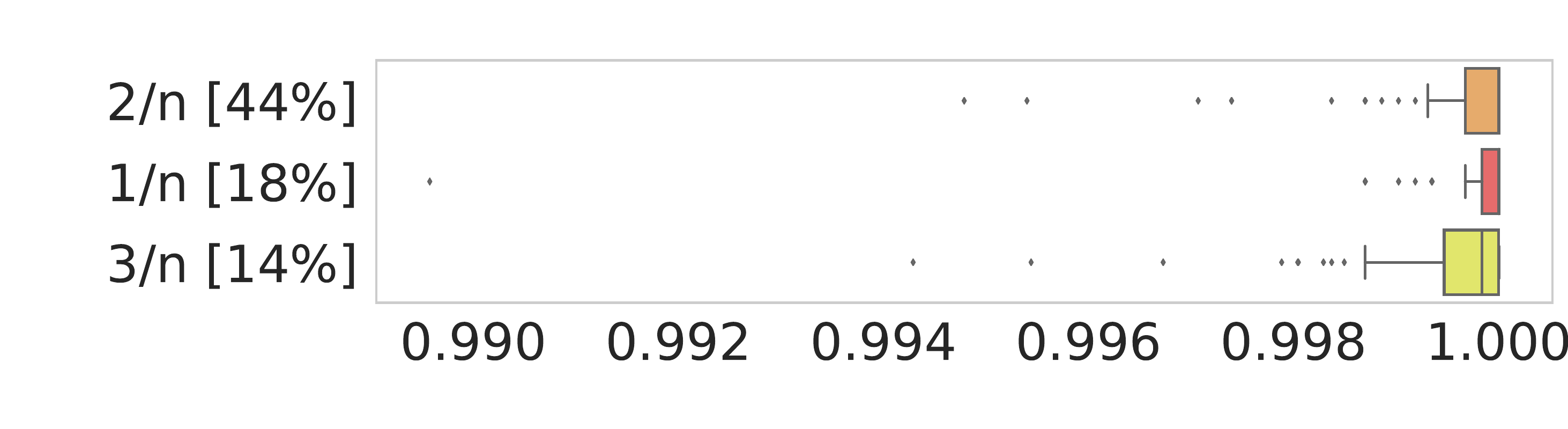}  & $n\ln(n)$*      & \includegraphics[width=.35\textwidth,clip,trim=70 0 0 0]{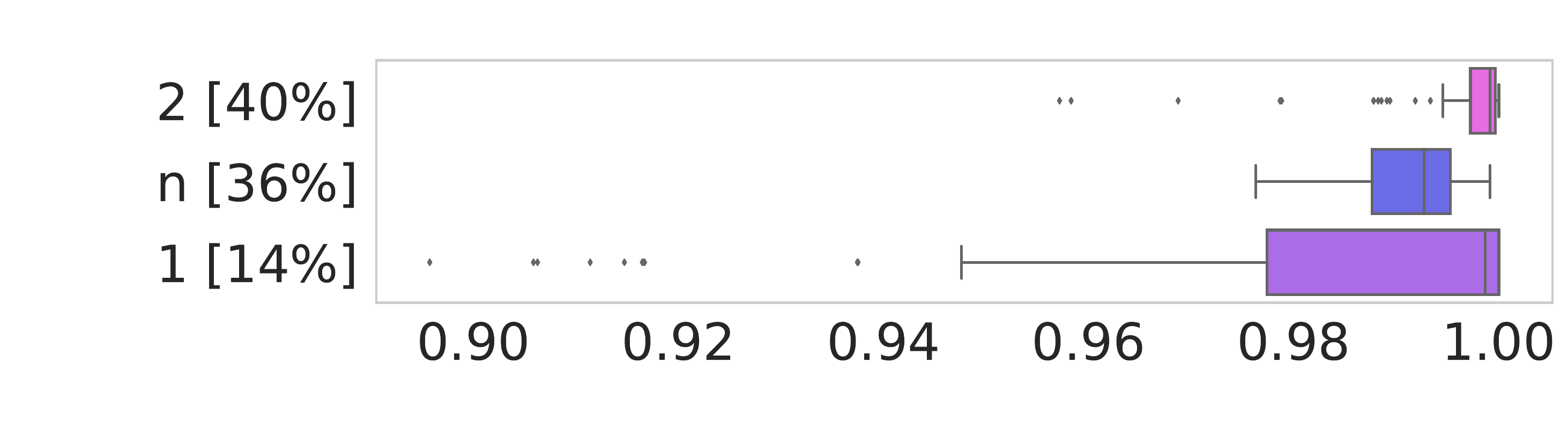}  \\ \cline{2-5}
  & $2en\ln(n)$**   & \includegraphics[width=.4\textwidth]{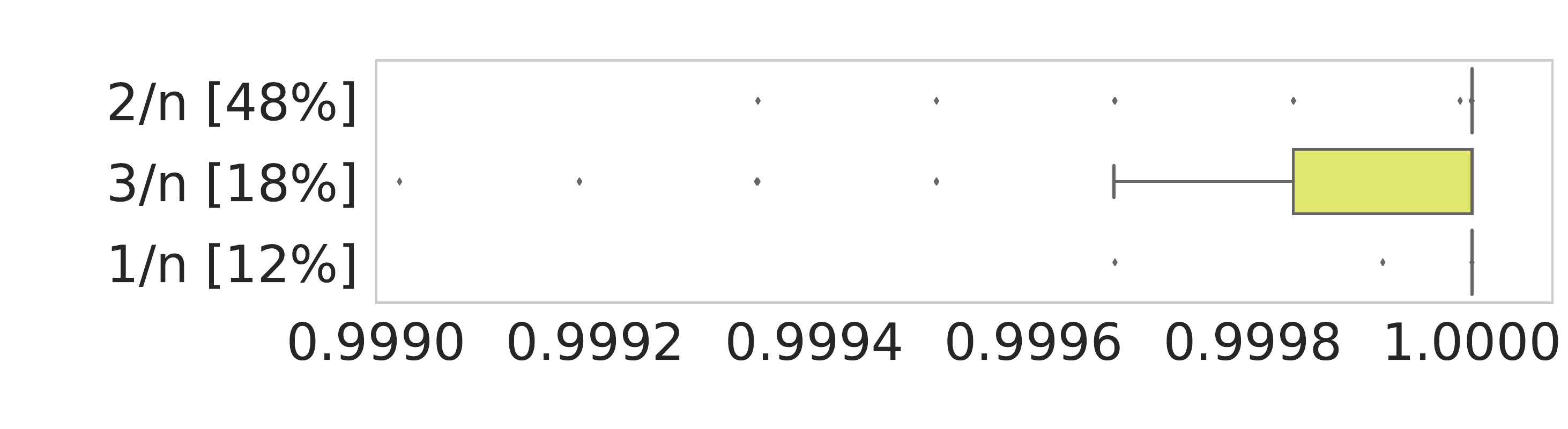}  & $2n\ln(n)$**     & \includegraphics[width=.35\textwidth,clip,trim=70 0 0 0]{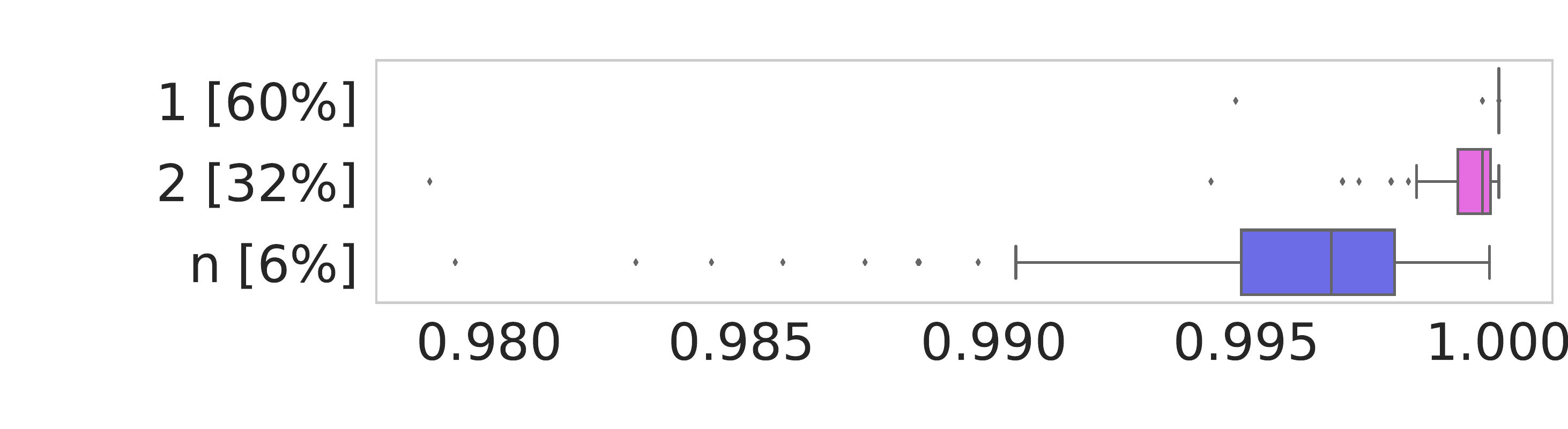}  \\ \hline
\multirow{3}{*}{\rotatebox[origin=c]{90}{\bf LeadingOnes}}
  & $0.5n^2$      & \includegraphics[width=.4\textwidth]{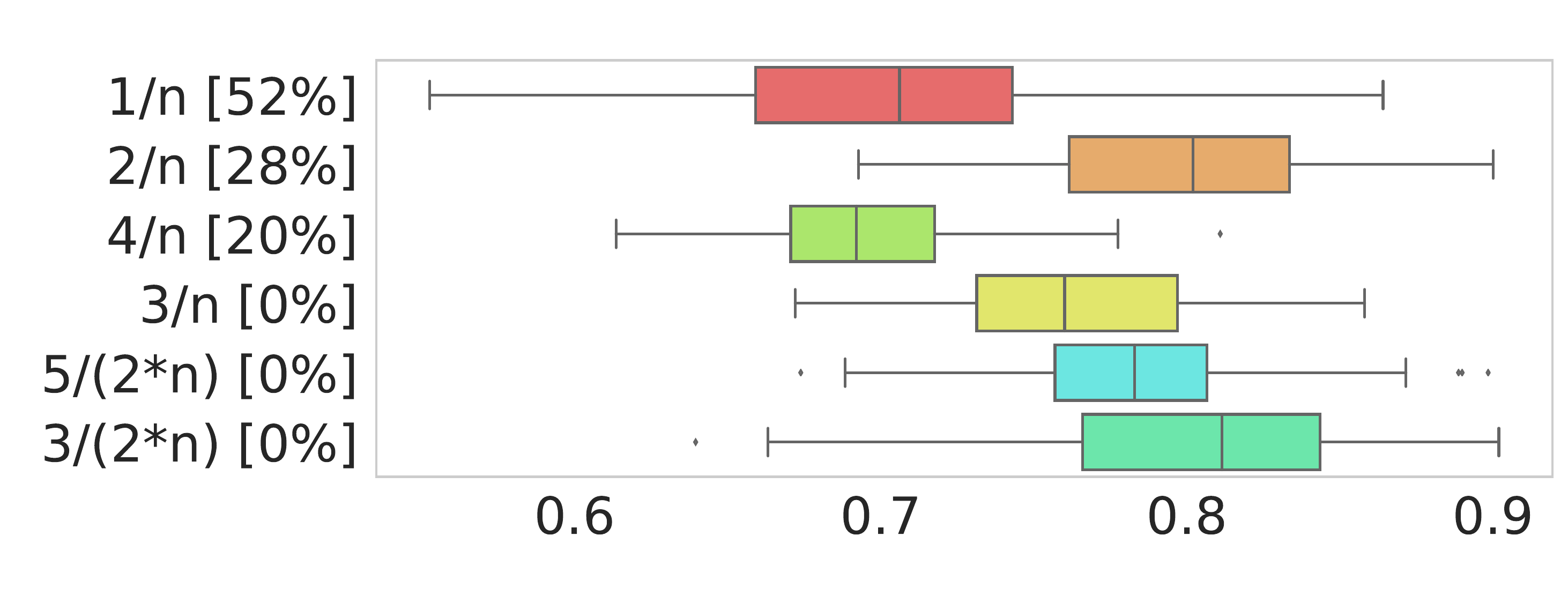}  & $0.5n^2$*       & \includegraphics[width=.35\textwidth,clip,trim=70 0 0 0]{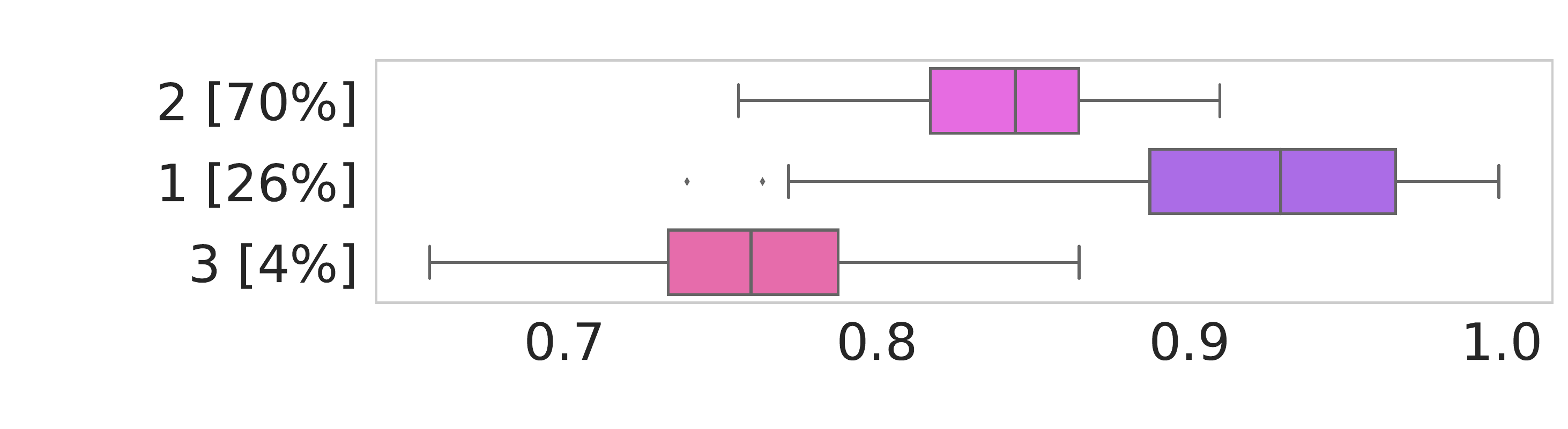}  \\ \cline{2-5}
  & $0.8n^2$**      & \includegraphics[width=.4\textwidth]{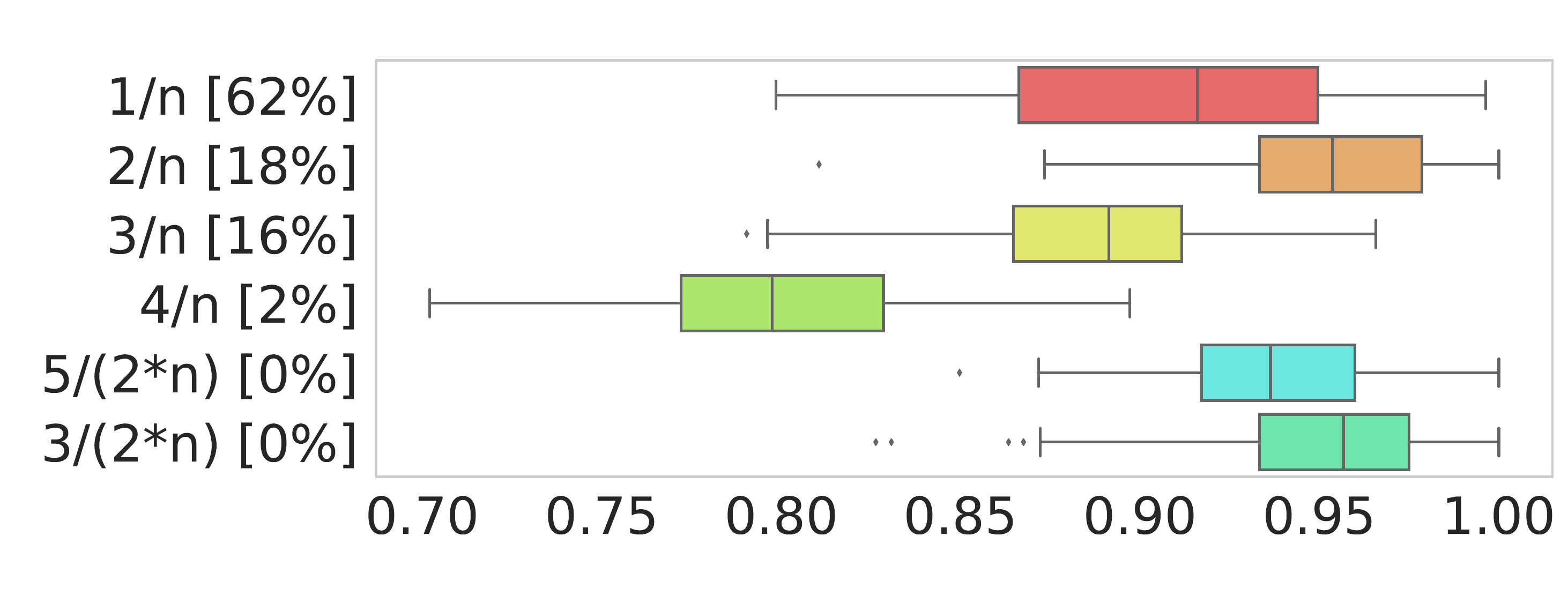}  & $0.75n^2$**      & \includegraphics[width=.35\textwidth,clip,trim=70 0 0 0]{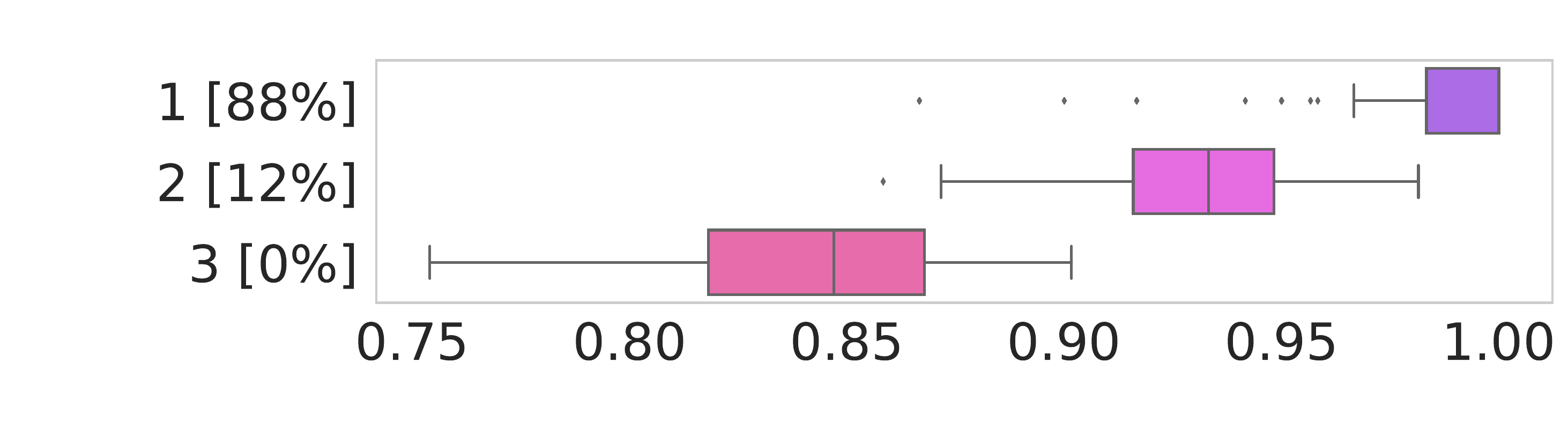}  \\ \cline{2-5}
  & $0.9n^2$**      & \includegraphics[width=.4\textwidth]{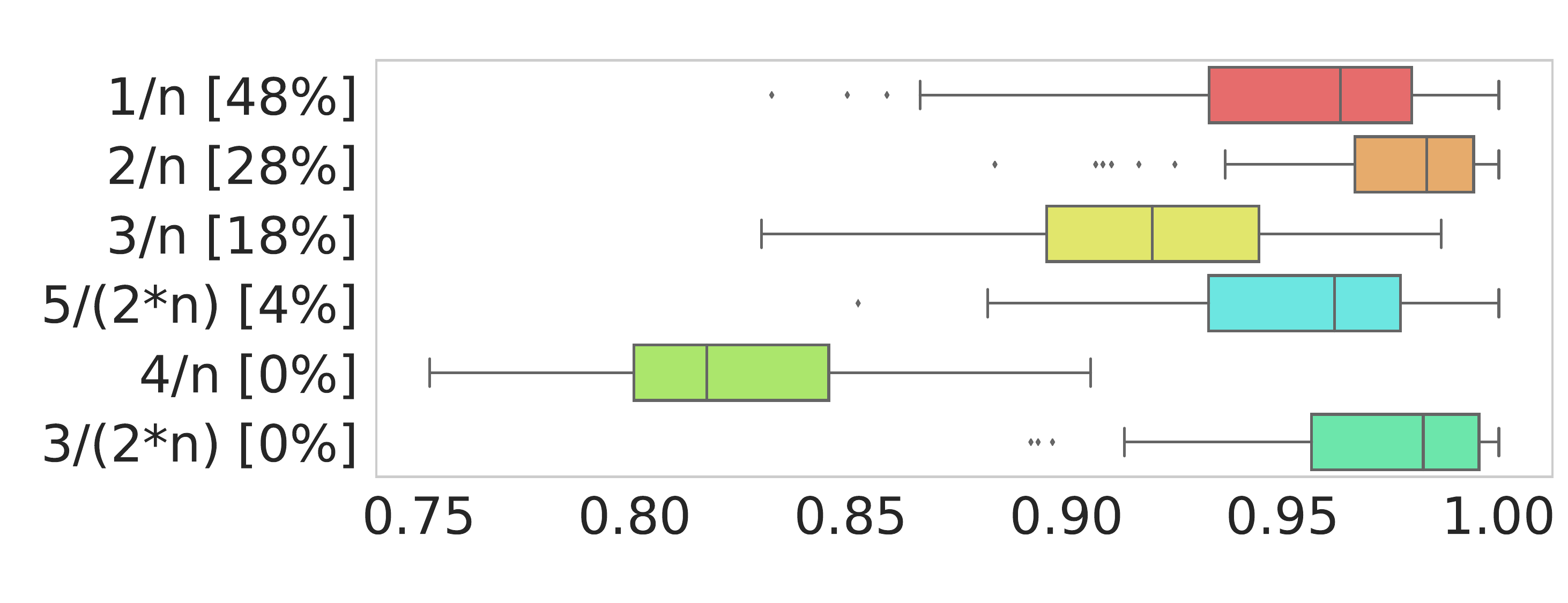}  &                &                                                              \\ \hline
\multirow{2}{*}{\rotatebox[origin=c]{90}{\bf Jump}}
  & $n^m$         & \includegraphics[width=.4\textwidth]{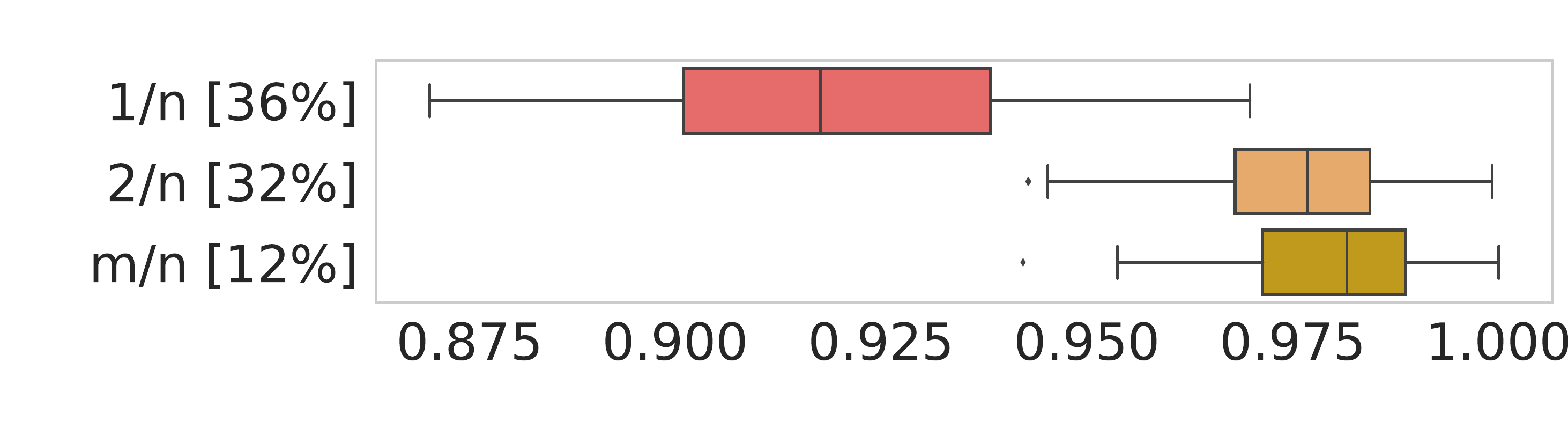} & $n^m$          & \includegraphics[width=.35\textwidth,clip,trim=45 0 0 0]{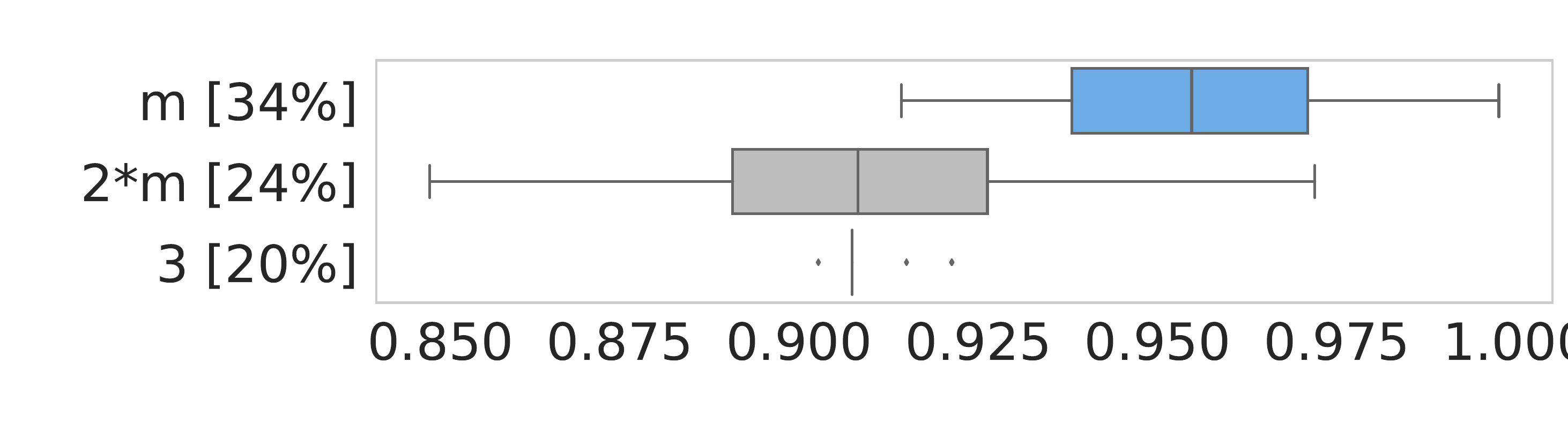} \\ \cline{2-5}
  & $en^m$**        & \includegraphics[width=.4\textwidth]{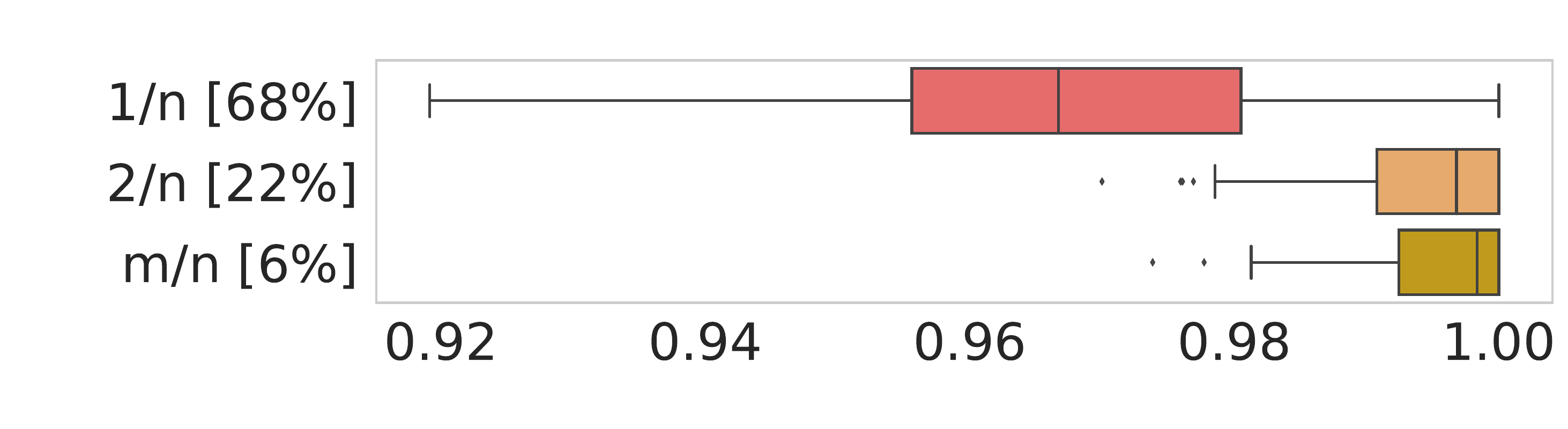} & $2n^m$         & \includegraphics[width=.35\textwidth,clip,trim=45 0 0 0]{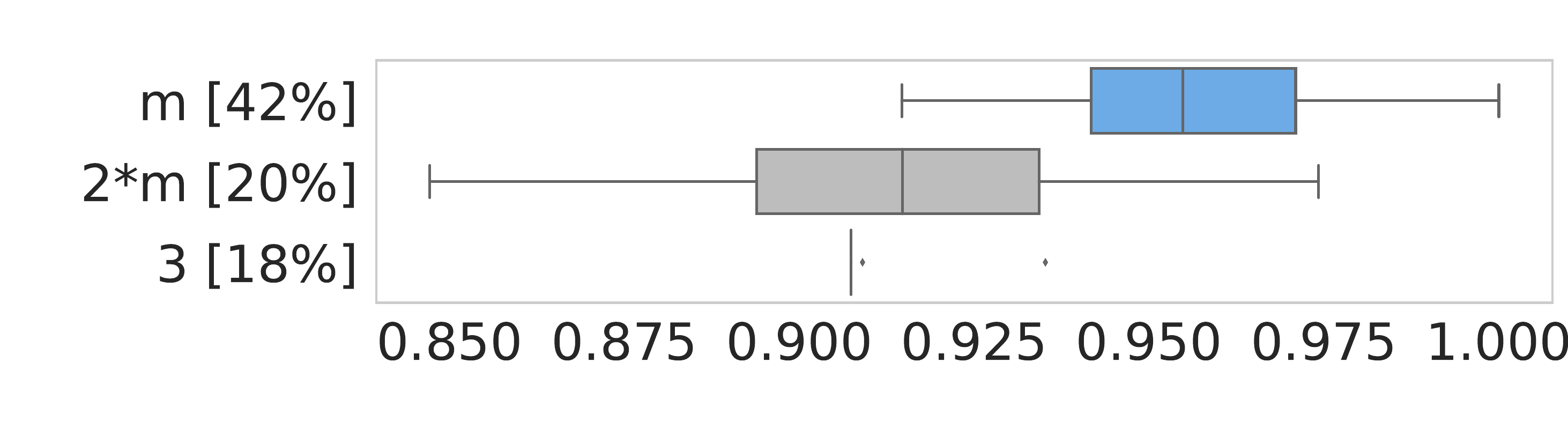} \\ 
\hline

\end{tabular}
\begin{tablenotes} [flushleft]
     \begin{small}
   \item[] $\dagger$ The $y$-axis show the best found expressions with its frequency between square brackets, and the $x$-axis represents the normalised fitness. 
     \end{small}
    \end{tablenotes}
\end{threeparttable}
\end{table}

\textit{Comparison amongst the top 3 configurations.}  When comparing the top 3 ranked configurations, we observe the following from Table~\ref{tab:results-boxplots} while we compare medians. 
\begin{itemize}
    \item OneMax: For (1+1)~EA, $\mu=1/n$, which
    ranked second for budgets $0.5e n\ln n$ and $e n\ln n$ and first for budget $2 e n\ln n$ performs better than $\mu=1/2*n$; 
    while for RLS, the expression $k=1$ appears at least on $94\%$, providing the best results;
    \item BinValue: $\mu=1/n$ represents $18\%$ on $e n\ln n$ for (1+1)~EA experiments, and a similar performance with $\mu=2/n$ and $\mu=3/n$; while on $0.5e n\ln n$ case the $\mu=1/n$ expression provides better results than $\mu=1/2$ and $\mu=1/3$; on the same way the expression $k=1$ corresponds to $60\%$ of the cases on RLS with the budget of $2 n\ln n$ with a better performance than $k=2$ and $k=n$;  
    \item LeadingOnes: $\mu=1/n$ is the most frequent expression among all considered budgets on (1+1)~EA and $\mu=2/n$ presents the best performance amongst the top 3 expressions for all budget cases;  $k=1$ represents $88\%$ on RLS cases with $0.75n^2$ iterations and performs better than $k=2$ and $k=3$ for both considered budgets.
    \item Jump: $\mu=2/n$ and $\mu=m/n$  present similar results for both budget cases; $\mu=1/n$ appears on $36\%$ and $68\%$ of the cases on (1+1)~EA on the considered budgets respectively, and performs worse than the other two $\mu$ configurations; for RLS experiments $k=m$ is the most frequent expression and performs better than $k=2*m$ and $k=3$.
\end{itemize}

\textit{Comparison of top 3 configurations against other parameter settings.} 
For a fair assessment of our results, we add to this comparison some expressions that were not ranked in the top 3. 
These are $\mu = i/n$ with $i \in \{1,3/2,2,5/2,3,4\}$ for (1+1)~EA$(\mu)$ for OneMax and LeadingOnes.
For readability purposes, the top $3$ expressions are complemented with $3$ of these additional expressions in the same order they are shown.
We can observe in Table~\ref{tab:results-boxplots} that these additional expressions present low frequencies, $\mu=3/n$ being the highest case with $12\%$ with the budget $e n\ln n$, while expressions $\mu=3/(2n)$ and $\mu=5/(2n)$ are the lowest cases among the considered budgets. Note that the frequencies do not necessarily sum up to $100\%$ as other expressions not reported here might have occurred.

\textit{Comparison with theoretical results.} As we have mentioned in the beginning of this section,
one should be careful when comparing theoretical results that have been derived either in terms of running time or in terms of asymptotic convergence analysis, as typically done in runtime analysis. It is well known that optimal parameter settings for concrete (typically, comparatively small) dimensions can be different from the asymptotically optimal ones~\cite{BuskulicD19,ChicanoSWA15}. We nevertheless see that the configurations that minimise the expected running times (again, in the classical, asymptotic sense) also show up in the top 3 ranked configurations. 
In Table~\ref{tab:results-boxplots}, we highlight the asymptotically optimal best possible running time by an asterisk *. Budgets exceeding this bound are marked by two asterisks **. As for the individual problems, we note the following: 

\begin{itemize}
    \item OneMax: It is interesting to note here that the performance is not monotonic in $k$, i.e., $k=2$ performs worse than $k=1$ and $k=3$. This is caused by a phenomenon described in~\cite[Section~4.3.1]{DoerrDY20}, which states that, regardless of the starting point, the expected progress is always maximised by an uneven mutation strength. MATE correctly identifies this and suggests uneven mutation strengths in almost all cases.
    \item BinValue: We observe that it is very difficult here  to distinguish the performance of the different configurations. This is in the nature of BinValues, as setting the first bit correctly already ensures 50\% of the optimal fitness values. We very drastically see this effect in the recommendation to use $k=n$ for the RLS cases. With this configuration, the algorithm evaluates only two points: the random initial point $x$ and its pairwise complement $\bar{x}$, regardless of the budget. As can be seen in Table~\ref{tab:results-boxplots}, the performance of this simple strategy is quite efficient, and hard to beat 
    \item LeadingOnes: As mentioned earlier, for the (1+1)~EA, the optimal mutation rate in terms of minimising the expected running time is around $\mu=1.59/n$. We see that $\mu=3/(2n)$, which did not show in the top 3 ranked configurations performs better 
    than any of the suggestions by MATE.
    \item Jump: as discussed, mutation rate $\mu=m/n$ minimises the expected optimisation time. MATE recognises it as a good configuration in some of the runs. However, we see that $\mu=2/n$, which equals $\mu=m/n$ for 5 out of our 12 training sets, shows comparable performance, and in the $en^m$ budget case even slightly better performance. 
\end{itemize}

\subsubsection{Evaluation Phase II}


\begin{table}[!h]
\scriptsize\vspace{-0.3cm}
\centering
\caption{Results for larger OneMax and LeadingOnes instances}
\label{tab:results-boxplots-large}
\renewcommand{\arraystretch}{1.0}
\begin{tabular}{|m{.25cm}|l|l|l|l|}
\hline
  & \multicolumn{2}{l|}{\textbf{1+1-EA}}  & \multicolumn{2}{l|}{\textbf{RLS}}  \\ \hline
  & \textbf{Budget}   & \textbf{Result}   & \textbf{Budget} & \textbf{Result}  \\ \hline

\multirow{3}{*}{\rotatebox[origin=c]{90}{\bf OneMax}}
  & $0.5en\ln(n)$ & \includegraphics[width=.4\textwidth]{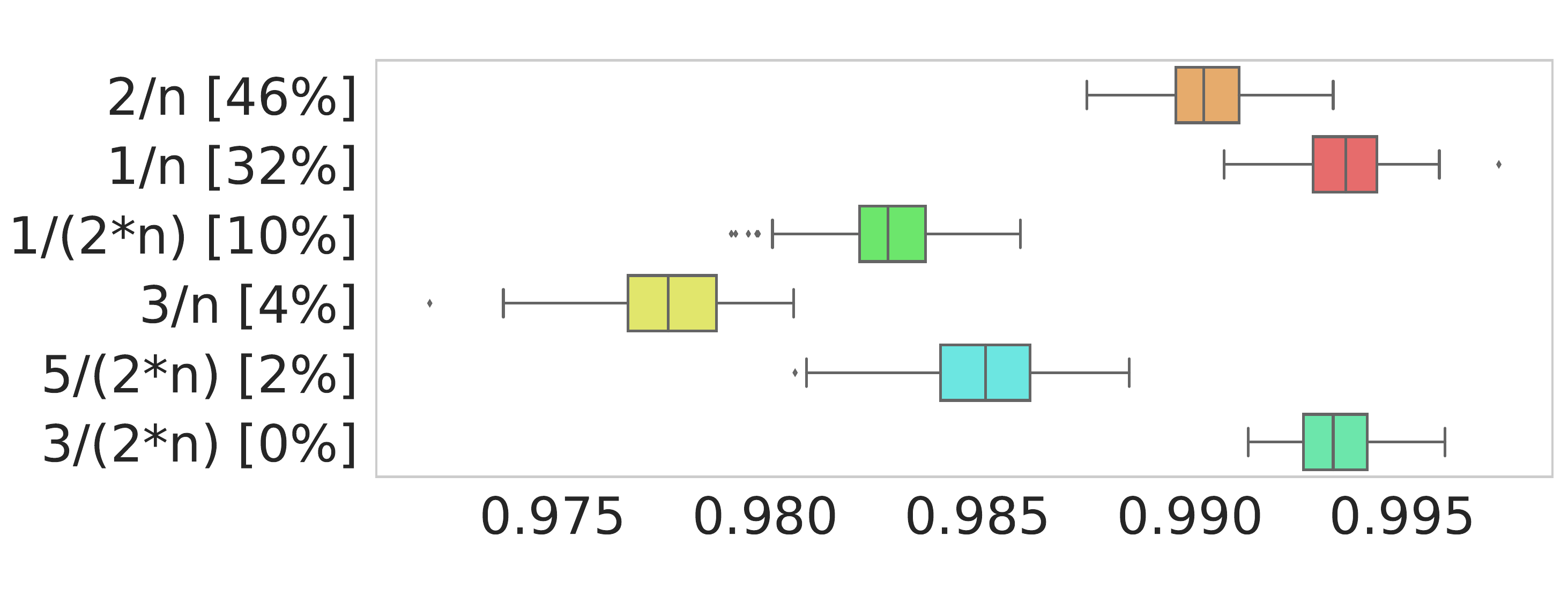}  & $n\ln(n)$*      & \includegraphics[width=.35\textwidth,clip,trim=70 0 0 0]{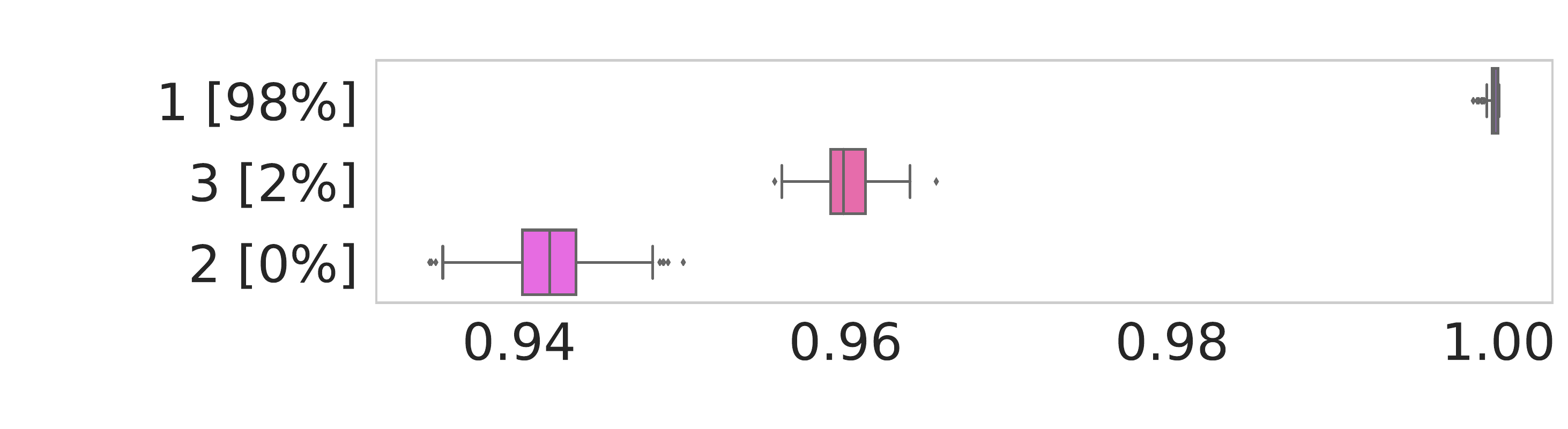}  \\ \cline{2-5}
  & $  e n\ln(n)$* & \includegraphics[width=.4\textwidth]{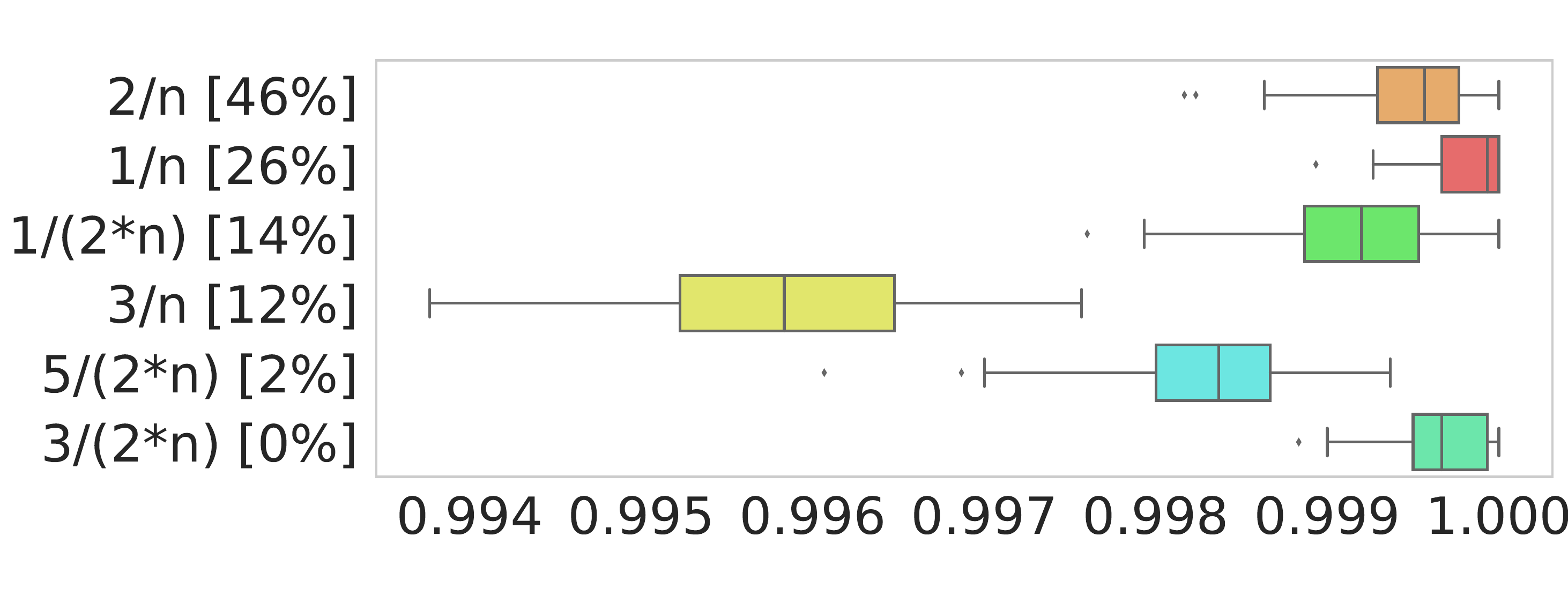}  & $2n\ln(n)$**     & \includegraphics[width=.35\textwidth,clip,trim=70 0 0 0]{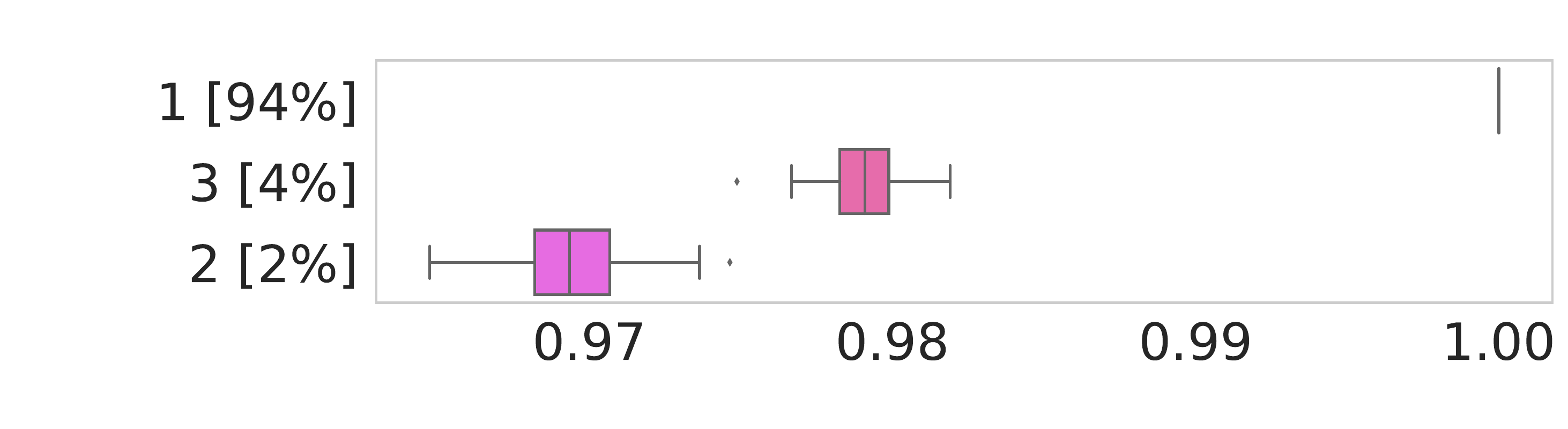}  \\ \cline{2-5}
  & $2 e n\ln(n)$** & \includegraphics[width=.4\textwidth]{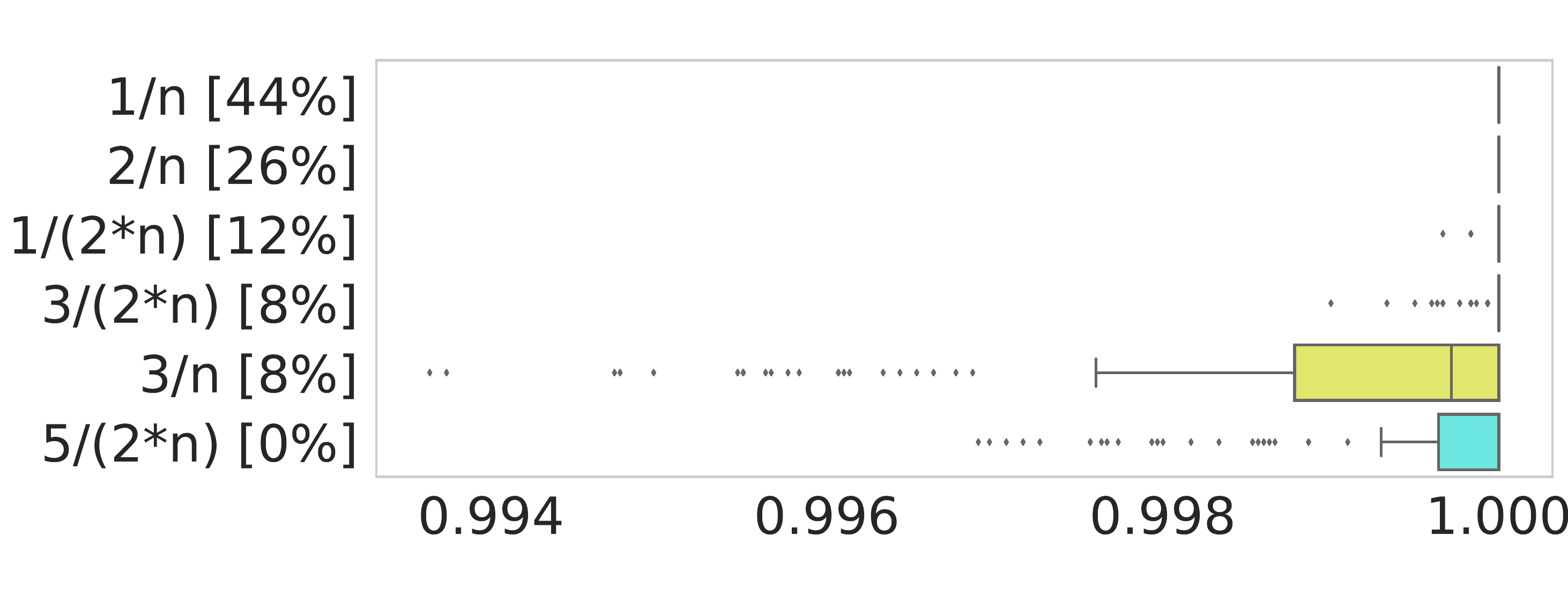}  &                &                                                              \\ \hline
\multirow{3}{*}{\rotatebox[origin=c]{90}{\bf LeadingOnes}}
  & $0.5n^2$      & \includegraphics[width=.4\textwidth]{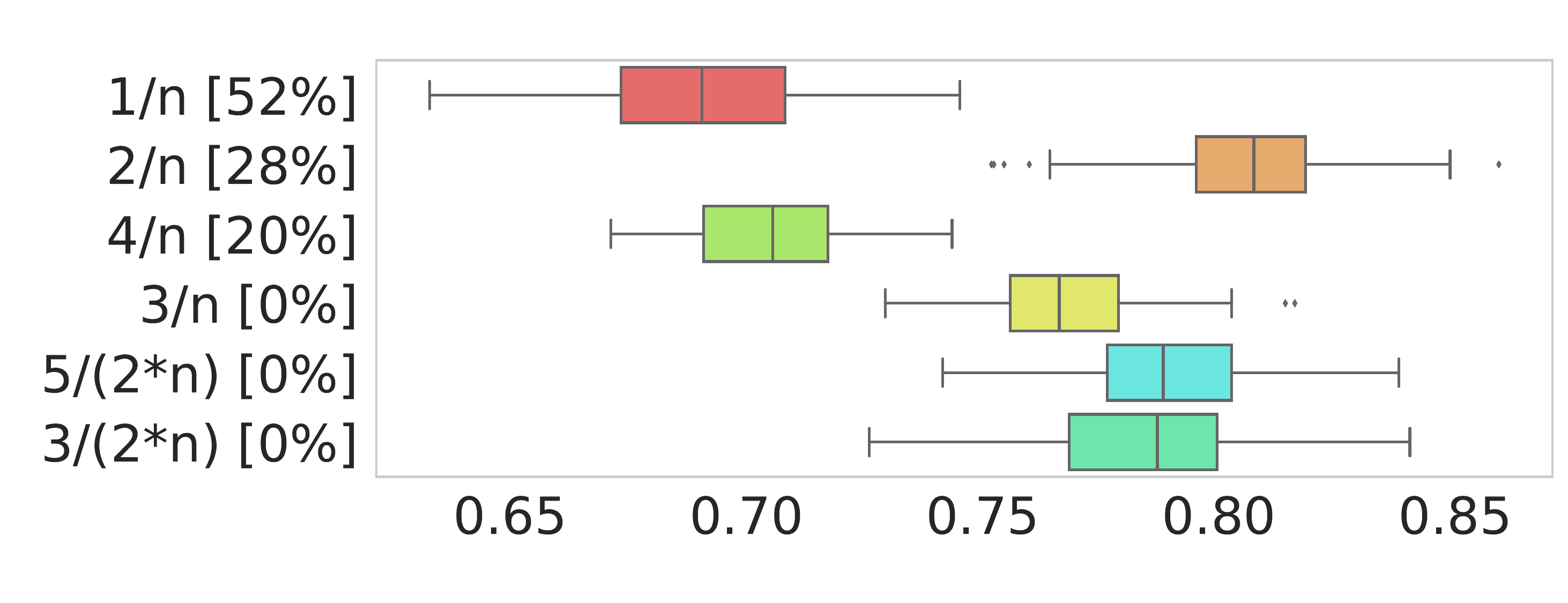}  & $0.5n^2$*       & \includegraphics[width=.35\textwidth,clip,trim=70 0 0 0]{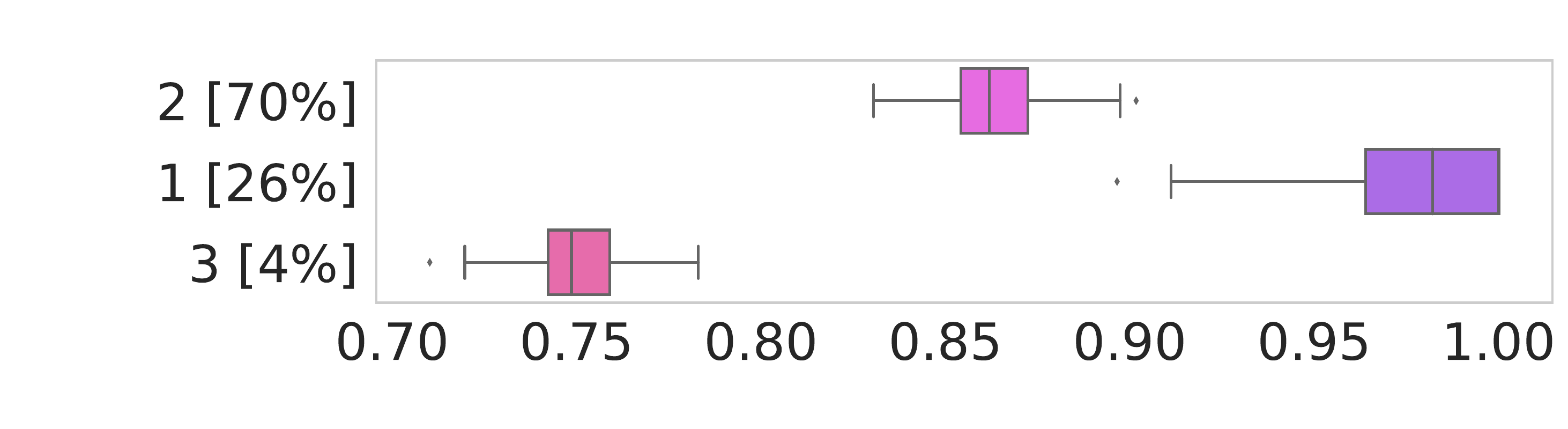}  \\ \cline{2-5}
  & $0.8n^2$**      & \includegraphics[width=.4\textwidth]{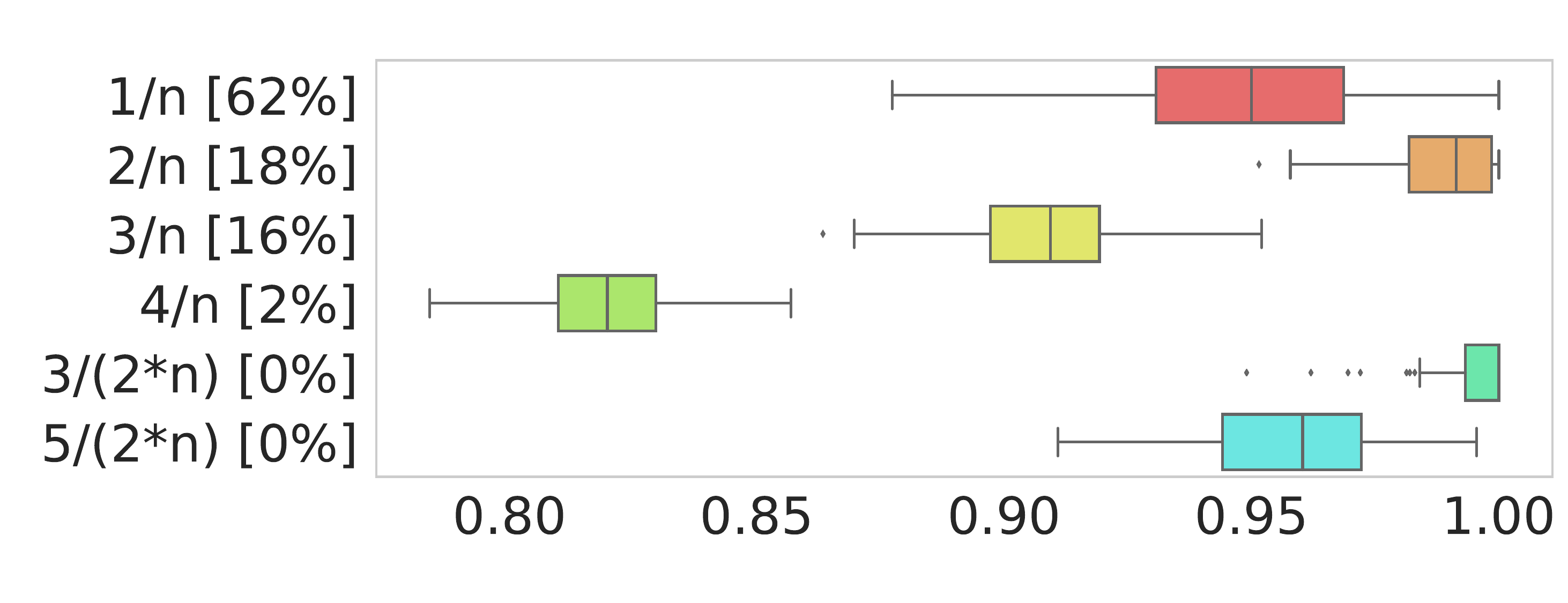}  & $0.75n^2$**      & \includegraphics[width=.35\textwidth,clip,trim=70 0 0 0]{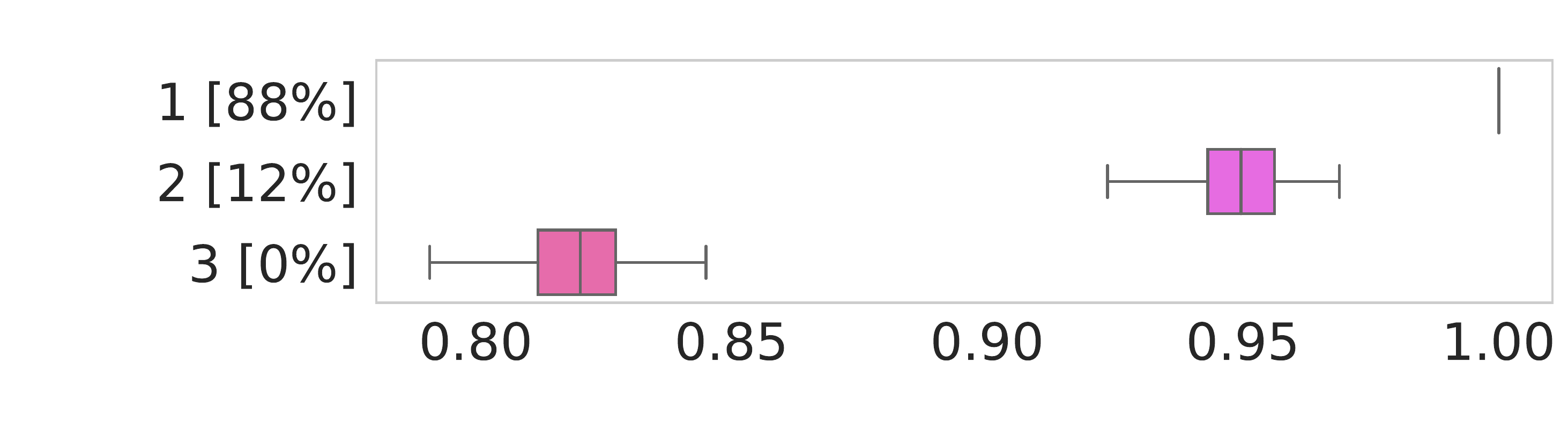}  \\ \cline{2-5}
  & $0.9n^2$**      & \includegraphics[width=.4\textwidth]{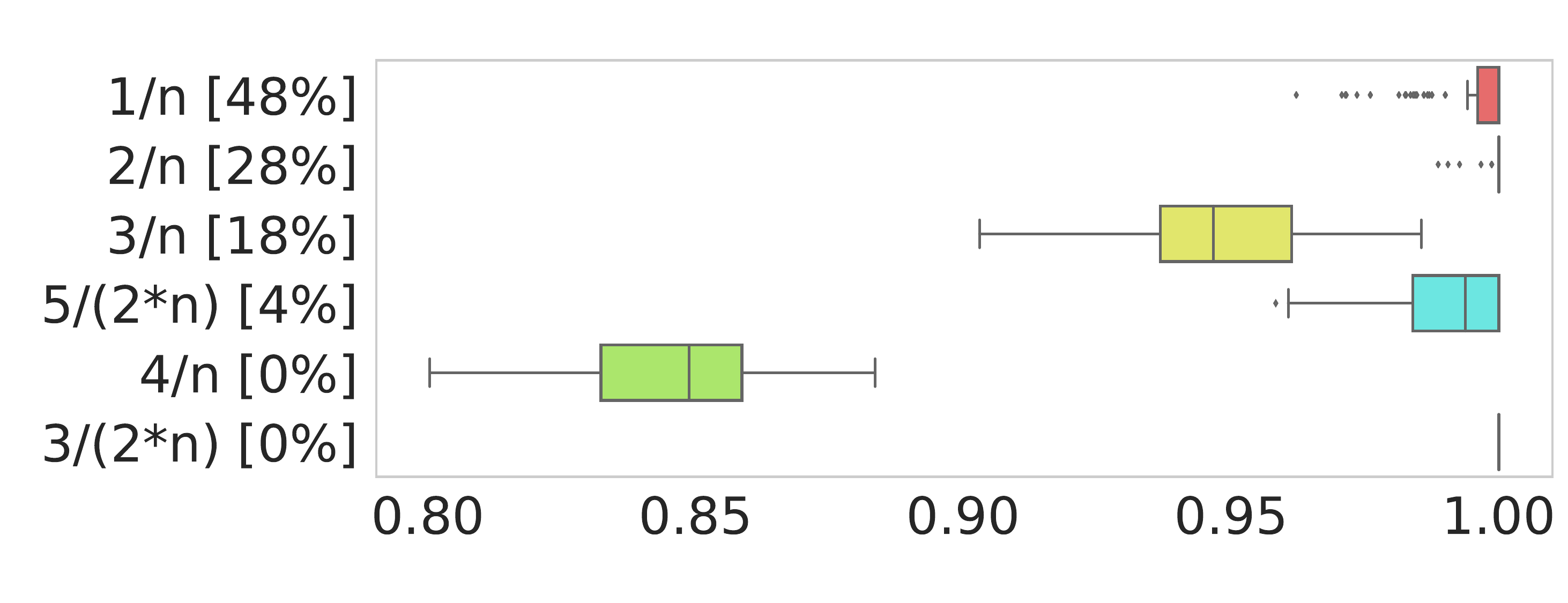}  &                &                                                              \\ \hline

\end{tabular}
\vspace{-0.2cm}
\end{table}

To properly assess the performance of MATE, we conducted experiments for OneMax and LeadingOnes instances of larger sizes that were not considered in the training phase. The goal of this experiment is to empirically demonstrate that our approach generalises well for large and unseen instances. These results are presented in Table~\ref{tab:results-boxplots-large} where $100$ runs were performed for OneMax with $n\in\{1000,2000,5000\}$ and LeadingOnes with $n\in\{750,1000\}$. We can observe the following:

\begin{itemize}
    \item There is less overlap amongst the confidence intervals especially for smaller budgets, which means there is a higher level of separability amongst the performances of the different expressions.
    \item By comparing these results with the ones from Table~\ref{tab:results-boxplots}, we can observe that the results of the top 3 expressions on large instances are statistically better in the majority of cases.
    \item OneMax: For (1+1)~EA, in contrast to the results in Table~\ref{tab:results-boxplots} where $\mu=1/n$ and $\mu=3/(2n)$ show a similar performance, here $\mu=1/n$ performs better than the other expressions.
    For RLS, the best performing expression is $k=1$, which was ranked first.
    \item LeadingOnes: For (1+1)~EA the best expressions are $\mu=2/n$, which was ranked second, and $\mu=3/(2n)$, which was not ranked among the top 3 expressions.
    For RLS, $k=1$, ranked first and second, is the best performing expression.
\end{itemize}

\subsection{Comparative study}

Herein, we compare the performance of MATE with irace and SMAC. The goal is to investigate the sensitivity of the obtained parameters on unseen instances. For a fair comparison, we run irace and SMAC with $2000$ maximum experiments (which we believe is equivalent to the $100$ GP generations with a population size of $20$ individuals in MATE) considering the training instances presented in Table \ref{tab:problems}. 
We report the best elite parameter values returned by irace (2 candidates), SMAC (1 candidate) and MATE (most frequent expressions) in the columns $\mu$ and $k$ in Table \ref{tab:comparison}, while the score (Eq.~\ref{eq:score}) is shown in column \textit{Score} with the standard deviation as a subscript. These parameter values are then applied over  $100$ runs performed for OneMax with $n\in\{1000,2000,5000\}$ and LeadingOnes with $n\in\{750,1000\}$.


{\renewcommand{\arraystretch}{1.6}
\begin{table}\centering\scriptsize
\caption{Results for MATE, irace and SMAC for OneMax and LeadingOnes instances.\label{tab:comparison}}
\arrayrulecolor{black}
\begin{tabular}{cl|l|l|l|l|l|l|l|r|l|l|l|l|l} 
\arrayrulecolor{black}\toprule
\multicolumn{1}{l}{} & \multicolumn{7}{c|}{\textbf{1+1-EA}} & \multicolumn{7}{c}{\textbf{RLS}} \\
\multicolumn{1}{l}{} &  & \multicolumn{2}{c|}{MATE} & \multicolumn{2}{c|}{irace} & \multicolumn{2}{c|}{SMAC} &  & \multicolumn{2}{c|}{MATE} & \multicolumn{2}{c|}{irace} & \multicolumn{2}{c}{SMAC}\\
\hline
\multicolumn{1}{l}{} & Budget & $\mu$ & Score & $\mu$  & Score & $\mu$ & Score & Budget & 
\multicolumn{1}{l|}{$k$} & Score & $k$ & Score & $k$ & Score \\ 
\midrule
\multirow{9}{*}{\rotatebox[origin=c]{90}{\bf OneMax}}      
& $\frac{enln(n)}{2}$ & $\frac{2}{n}  $ & $0.99_{0.001}$ & $0.258$  & $0.57_{0.002}$ & $0.009$ & $0.8_{0.003}$ & $nln(n)$ & $1$ & $1_0$          & \multicolumn{1}{r|}{1} & $1_0 $ & \multicolumn{1}{r|}{1} & $1_0$ \\
&              & $\frac{1}{n}  $ & $0.99_{0.001}$ & $0.216$  & $0.58_{0.002}$ &         &               &          & $3$ & $0.96_{0.002}$ &                        &        &                        &       \\
&              & $\frac{1}{2n} $ & $0.98_{0.002}$ &          &                &         &               &          & $2$ & $0.94_{0.003}$ &                        &        &                        &       \\
\cline{2-15}
& $enln(n)$ & $\frac{1}{n}$  & $1_0$ & $0.009$  & $0.82_{0.002}$ & $0.016$ & $0.76_{0.003}$ & $2nln(n)$ & $1$ & $1_0$          & \multicolumn{1}{r|}{1} & $1_0$ & \multicolumn{1}{r|}{1} & $1_0$ \\
&           & $\frac{2}{n}$  & $1_0$ & $0.013$  & $0.79_{0.003}$ &         &                &           & $3$ & $0.98_{0.001}$ &                        &       &                        &       \\
&           & $\frac{1}{2n}$ & $1_0$ &          &                &         &                &           & $2$ & $0.97_{0.002}$ &                        &       &                        &       \\ 
\cline{2-15}
& $2enln(n)$ & $\frac{2}{n}$  & $1_0$ & $0.594$ & $0.54_{0.002}$ & 0.008 & $0.86_{0.002}$ &  & \multicolumn{1}{l|}{} &  &  &  &  & \\
&            & $\frac{1}{n}$  & $1_0$ & $0.589$ & $0.54_{0.002}$ &       &                &  & \multicolumn{1}{l|}{} &  &  &  &  & \\
&            & $\frac{1}{2n}$ & $1_0$ &         &                &       &                &  & \multicolumn{1}{l|}{} &  &  &  &  & \\
\midrule
\multirow{9}{*}{\rotatebox[origin=c]{90}{\bf LeadingOnes}} 
& $0.5n^2$ & $\frac{1}{n}$ & $0.7_{0.025}$  & $0.430$ & $0.03_{0.002}$ & $0.024$ & $0.29_{0.007}$ & $0.5n^2$  & $2$ & $0.86_{0.014}$ & \multicolumn{1}{r|}{1} & $0.98_{0.026} $ & \multicolumn{1}{r|}{1} & $0.98_{0.02}$ \\
&          & $\frac{2}{n}$ & $0.8_{0.021}$  & $0.409$ & $0.03_{0.002}$ &         &                &           & $1$ & $0.97_{0.027}$ & \multicolumn{1}{r|}{5} & $0.61_{0.01}  $ &                        &               \\
&          & $\frac{4}{n}$ & $0.71_{0.015}$ &         &                &         &                &           & $3$ & $0.75_{0.013}$ &                        &                 &                        &               \\ 
\cline{2-15}
& $0.8n^2$ & $\frac{1}{n}$ & $0.95_{0.023}$ & $0.255$ & $0.05_{0.002}$ & $0.005$ & $0.83_{0.017}$ & $0.75n^2$ & $1$ & $1_0$          & \multicolumn{1}{r|}{1} & $1_0$ & \multicolumn{1}{r|}{1} & $1_0$       \\
&          & $\frac{2}{n}$ & $0.99_{0.012}$ & $0.258$ & $0.05_{0.002}$ &         &                &           & $2$ & $0.95_{0.009}$ &                        &       &                        &             \\
&          & $\frac{3}{n}$ & $0.91_{0.018}$ &         &                &         &                &           & $3$ & $0.82_{0.01}$  &                        &       &                        &             \\ 
\cline{2-15}
& $0.9n^2$ & $\frac{1}{n}$ & $0.99_{0.011}$ & $0.158$ & $0.07_{0.003}$ & $0.006$ & $0.75_{0.013}$ &  & \multicolumn{1}{l|}{} &  &  &  &  &  \\
&          & $\frac{2}{n}$ & $1_0$          & $0.153$ & $0.07_{0.006}$ &         &                &  & \multicolumn{1}{l|}{} &  &  &  &  &  \\
&          & $\frac{3}{n}$ & $0.95_{0.014}$ &         &                &         &                &  & \multicolumn{1}{l|}{} &  &  &  &  &  \\
\bottomrule
\end{tabular}
\arrayrulecolor{black}
\end{table}
}

Table~\ref{tab:comparison} shows that MATE significantly outperforms irace and SMAC for (1+1)~EA. On the other hand, the three methods show a similar performance on RLS.
This is due to the fact that the parameter $\mu$ in (1+1)~EA is highly sensitive to the problem feature $n$. In contrast, the parameter $k$ in RLS is independent from $n$ and its best value ($k=1$) was identified by the three methods for both OneMax and LeadingOnes.

\section{Conclusions and Future Directions}
\label{sec:sect5}

With this article, we have presented MATE as a model-based algorithm tuning engine: its human-readable models map instance features to algorithm parameters. Our experiments showed that MATE can find known asymptotic relationships between the feature values and algorithm parameters. We also compared the performance of MATE with iRace and SMAC investigating the sensitivity of the obtained parameters on unseen instances of larger size. With its scalable models, MATE performed best. It is worth noting that MATE can be a useful guideline tool for theory researchers due to its white-box nature, similarly to how results in~\cite{DoerrW18} inspired the analysis of a  generalised one-fifth success rule in~\cite{DoerrDL19}. But MATE can also be extended to be used as a practical toolbox for feature-based algorithm configuration.



In the future, we intend to explore, among other, the following three avenues. 
First, the design of MATE itself will be subject to extensions, e.g. to better handle performance differences between instances via ranks or racing. 
Second, while our proof-of-concept study here was motivated by theoretical insights, we will investigate more realistic problems for which instance features are readily available, such as the travelling salesperson problem and the assignment problem.
Third, we will investigate approaches to extend MATE to handle multiple parameters to demonstrate its ability to tune more sophisticated algorithms.



\section*{Acknowledgements} 
M. Martins acknowledges CNPq (Brazil Government).
M. Wagner acknowledges the ARC Discovery Early Career Researcher Award DE160100850. 
C. Doerr acknowledges support from the Paris Ile-de-France Region. 
Experiments were performed on the AAU's CLAUDIA compute cloud platform.

\bibliographystyle{splncs04}
\bibliography{main}

\end{document}